\documentclass[letterpaper]{article}
\usepackage{arxiv}

\def\makeenmark{$^{\theenmark}$}
\def\enoteformat{\rightskip0pt\leftskip0pt\parindent=1.75em
  \leavevmode\llap{\theenmark.\enskip}}

\usepackage{blkarray}
\usepackage[thinc]{esdiff}
\usepackage{bm}
\usepackage{bbm}
\usepackage{graphicx}
\usepackage{url}

\usepackage{subfig}
\usepackage{subfloat}
\usepackage{xcolor}
\usepackage{booktabs}
\usepackage{enumitem}
\usepackage{amsmath,amsfonts,amssymb,amsthm}
\usepackage{cleveref}

\usepackage{tikz}
\usepackage{pgfplots}
\pgfplotsset{compat=1.11}
\usetikzlibrary{calc}

\usepackage{natbib}
 \bibpunct[, ]{(}{)}{,}{a}{}{,}%

\newcommand{\bla}{\color{black}}
\newcommand{\gra}{\color{gray}}

\newcommand{\Var}{\mathrm{Var}}
\newcommand{\Cov}{\mathrm{Cov}}
\newcommand{\IPS}{\mathrm{IPS}}
\newcommand{\CIPS}{\mathrm{CIPS}}

\newtheorem{definition}{Definition}

\DeclareMathOperator*{\argmax}{arg\,max}
\DeclareMathOperator*{\argmin}{arg\,min}
 
\newtheorem{theorem}{Theorem}
\newtheorem{corollary}{Corollary}[theorem]
\newtheorem{lemma}[theorem]{Lemma}
\newtheorem{assumption}{Assumption}
\newtheorem{example}{Example}

\newtheorem{proposition}{Proposition}

\newtheorem{repeattheorem}{Theorem}

\theoremstyle{EXkey}

\newtheorem{repeatexample}{Example}

\counterwithin*{footnote}{section}
\newcommand{\Halmos}{\ensuremath{\blacksquare}}

\title{Loss Functions for Discrete Contextual Pricing with Observational Data}
\author{
  Max Biggs\thanks{Alphabetical Order} \\
  Darden School of Business, University of Virginia \\
  \texttt{mbiggs@darden virginia.edu} \\
   \And
  Ruijiang Gao\footnotemark[1] \\
  University of Texas at Austin \\
  \texttt{ruijiang@utexas.edu} \\
  \And 
  Wei Sun\footnotemark[1] \\
  IBM Watson \\
  \texttt{sunw@us.ibm.com}}
\begin{document}

\maketitle

\begin{abstract}
    We study a pricing setting where each customer is offered a contextualized price based on customer and/or product features. 
Often only historical sales data are available, so we observe whether a  customer purchased a product at the price prescribed rather than the customer’s true valuation. 
Such observational data are influenced by historical pricing policies, which introduce difficulties in evaluating the effectiveness of future policies. 
The goal of this paper is to formulate loss functions that can be used for evaluating pricing policies directly from observational data, rather than going through an intermediate demand estimation stage, which may suffer from bias. To achieve this, we adapt ideas from machine learning with corrupted labels, where we consider each observed  purchase decision as a known probabilistic transformation of the customer's valuation. From this transformation, we derive a class of  unbiased loss functions. Within this class, we identify minimum variance estimators and estimators robust to poor demand estimation. 
Furthermore, we show that for contextual pricing, estimators popular in the off-policy evaluation literature fall within this class of loss functions. We offer managerial insights into scenarios under which these estimators are effective. 
\end{abstract}

\section{Introduction}

With increasing amounts of data gathered about customers %
and recent advances in machine learning, firms are looking for ways to leverage these data to improve their pricing policies. %
Contextual pricing aims to prescribe prices as a function of the data describing the customers or products. 
It has attracted the attention of many companies, 
including Airbnb \citep{ye2018customized}, Stubhub \citep{alley2019pricing}, and  ZipRecruiter \citep{dube2017scalable}. 
In addition to increased revenue, the benefits of contextual pricing include large-scale pricing automation.

In the setting we study, we only have access to observational data, meaning we only observe whether a customer purchased a product at the price they were offered, and we do not observe  counterfactual outcomes, that is, whether the customer would purchase if a different price had been offered. 
This lack of counterfactuals in observed data is known as the fundamental problem of causal inference \citep{holland1986statistics}. 
If a company has already attempted some form of contextual pricing, the observed data will likely be influenced by historical pricing policies and be imbalanced, with some prices being much more common for some customers than others. This can make it difficult to estimate customers' responses to uncommon prices \citep{shalit2017estimating}. The gold standard for data collection with potential outcomes is through randomized controlled trials (RCT) \citep{li2015value}, where treatments (i.e., prices) are assigned at random. %
As RCTs are often costly, time-consuming, and limited in size,  it is challenging to rely on them to evaluate customers' heterogeneous pricing preferences.  
Meanwhile, observational data from  past pricing decisions are readily abundant, offering the potential to improve upon pricing policies. %

A common approach used for contextual pricing is to first estimate a demand function from  data, and subsequently use it to determine the optimal price \citep{chen2015statistical, ferreira2016analytics, dube2017scalable, baardman2018detecting, alley2019pricing, biggs2021model}. More specifically, the demand function estimates the probability of each customer purchasing a product as a function of the offered price and customer/product features. 
One can then construct an estimator for the expected revenue (i.e., price multiplied by the purchase probability) and prescribe the price that maximizes this quantity. 
This approach is known as ``predict-then-optimize''  \citep{elmachtoub2017smart} or ``direct comparison/direct method''  in the causal inference literature~\citep{kallus2020efficient}. 
Apart from this method being inherently indirect, as the estimand of interest for the sequential approach is demand instead of price, 
its main drawback  is that the revenue estimation relies heavily on the estimated demand being an accurate representation of the ground truth. 
In practice, the estimated demand routinely suffers from bias. 
This may occur due to model misspecification \citep{jagabathula2017nonparametric}. For instance, a seller may intentionally select a model from a restricted class of functions for interpretability and ease of solving the downstream revenue optimization problem. 
Another potential source of bias common with machine learning models is introduced by regularization, which is intended to mitigate the over-fitting of the predictive model.
Furthermore, adopting modern deep neural networks or gradient-boosted trees can lead to poor uncertainty calibration,  whereby the probabilities of sale estimates are not accurate, even though the model may have high accuracy in predicting the true labels~\citep{niculescu2005obtaining, guo2017calibration}.

Due to the aforementioned issues,  the  estimated revenue that utilizes the demand function as an input may also be biased. 
In this work, we propose \textit{unbiased} loss functions for pricing with observational data without imposing restrictive structural assumptions on demand. These loss functions can be directly optimized to find a pricing policy. %
In the contextual pricing setting, we aim to measure the performance of a pricing policy, which is a mapping from customer/product features to a proposed discrete price. However, unlike in typical supervised learning settings, we do not observe the ideal price to charge each customer (their valuation or willingness to pay), which could be considered the true labels we are trying to learn. 

When valuation data are available, loss functions for pricing have been proposed by \cite{mohri2014learning}. Under this loss function, if the customer is offered a price below their valuation, a sale occurs and the reward accrued is the price prescribed. 
If the customer is offered a price above their valuation, no sale occurs and the reward gained is 0. In practice, valuation data is rarely available. 
To adapt this loss function to our pricing setting where valuations are not observed, we apply the theory of corrupted labels \citep{cour2011learning, natarajan2013learning, van2017theory, kallus2019classifying}. Under this framework, the outcomes that we would ideally observe (the \textit{clean labels}) are probabilistically transformed, resulting in \textit{corrupted labels} that are actually observed. %
More specifically, in the contextual pricing setting, we consider each observed outcome, i.e., whether a customer purchased at a prescribed price (the \textit{corrupted label}), as a known probabilistic transformation of the customer's valuation (the \textit{clean label}). With a reconstruction of the inverse of this probabilistic transformation, we are able to transform the valuation loss function to form a class of \textit{unbiased} loss functions that can be used to assess the performance of %
a future policy using observational data. In more detail, we make the following contributions:

\begin{itemize}
\item We introduce a class of \textit{unbiased} loss functions that can be used for contextual pricing policy evaluation, by bridging the gap between the theory of learning from corrupted labels and offline policy evaluation. %
Within this class of unbiased loss functions, we identify the loss function with minimum conditional variance in the scenario where we have access to the true demand function, or an accurate estimate thereof. We also provide generalization bounds for this setting to show how our approaches perform from a finite sample of observations.
\item For scenarios where accurate  demand estimation is challenging, we present a robust approach in which the loss function is chosen to minimize the conditional variance for all demand functions within an uncertainty set, which can be calibrated to capture the uncertainty in the estimates of the demand function. 
Furthermore, we are able to find a closed-form solution to this robust optimization problem, so that the loss function can be calculated efficiently. We also provide insights into which valuation distributions are challenging for pricing in this setting. 
\item We connect this class of loss functions to approaches from the causal inference literature. In particular, we show that when adapted to the contextual pricing setting, the inverse propensity scoring \citep{rosenbaum1983central, rosenbaum1987model, beygelzimer2009offset,li2011unbiased} and doubly robust \citep{robins1994estimation, dudik2011doubly, zhou2018offline} estimators are specific examples of loss functions in this class. 
To the best of our knowledge, this link between learning with corrupted labels and causal inference has not been established before. 
Furthermore, we are able to provide insights into when inverse propensity scoring methods are likely to be effective in this contextual pricing setting. Specifically, we show that inverse propensity scoring is a minimum variance loss function when all customers have the lowest possible valuation, and that it performs well when the purchase probability is low. %
We also introduce another estimator that has the minimum variance when all customers have the highest valuation and performs well when purchase probabilities are high. 

\item Finally, we provide extensive numerical experiments on both synthetic and real-world datasets to investigate the performance of the proposed estimators. %
As demand estimation may  be inaccurate, we demonstrate the superior performance of the proposed locally robust estimator for varying degrees of demand estimation quality.  
\end{itemize}

\bla

\section{Related Literature}

In recent years, there has been significant interest in learning contextual pricing algorithms, which incorporate customer and product features to pricing decisions. A common approach is the ``predict-then-optimize'' framework, where a demand function is estimated and then optimized to find the optimal pricing policy \citep{chen2015statistical, ferreira2016analytics, dube2017scalable, baardman2018detecting, alley2019pricing, biggs2021model,subramanian2022constrained}.
Some existing approaches assume access to customer valuation data or its distribution  \citep{mohri2014learning,  devanur2016sample, medina2017revenue, huchette2020contextual, elmachtoub2018value}. Meanwhile, some recent approaches study a similar setting that utilizes  data that record whether customers made  purchases at the prices prescribed \citep{ye2018customized,chen2021model}, but these estimators are generally not unbiased since the bias introduced by the historical pricing policy is not taken into account.
There are also many dynamic pricing settings where the demand function or pricing policy is learned over time \citep{bitran2003overview,vulcano2002optimal,gallego2006dynamic, araman2009dynamic,feng2010integrating, broder2012dynamic,harrison2012bayesian, javanmard2016dynamic, cohen2016feature, qiang2016dynamic,cheung2017dynamic,besbes2018dynamic, nambiar2019dynamic, ban2020personalized,calmon2021revenue,keskin2021data}. In such an online setting, the goal of the pricing policy is to ensure high long-term profits by balancing exploration and exploitation. 
However, online experiments are often costly and difficult to deploy, so we focus on the problem of finding and evaluating a pricing policy using observational data, which typically is readily available.

There has also been recent work on robust pricing \citep{bergemann2011robust,araman2011revenue,kos2015selling, carrasco2018optimal,  chen2019distribution,cao2019dynamic, cohen2021simple}. \cite{cohen2021simple} provide bounds when only the support of the valuation distribution is known; \cite{chen2019distribution} use the mean and variance of the valuation distribution; \cite{bergemann2011robust} use a neighborhood containing the true valuation distribution. Much of this work is aimed at pricing for new products where there is no, or very little, data on historical sales. In contrast, we consider a different ``robust'' scenario where  %
plug-in estimates of the demand may be inaccurate. 
There is also recent work on the identifiability in pricing \citep{bertsimas2016power}. Comprehensive reviews of the pricing literature can be found in \cite{gallego2019revenue} and \cite{chen2023data}.  

An alternative, but so far less well-known approach to contextual pricing is to apply off-policy evaluation and optimization techniques from the causal inference community. These include inverse propensity scoring (IPS) estimators \citep{rosenbaum1983central, rosenbaum1987model, beygelzimer2009offset,li2011unbiased}, where the reward from a policy is estimated by inversely weighting each reward by the treatment assignment probability, which is known as the propensity score; direct method \citep{qian2011performance,johansson2016learning,shalit2017estimating,kunzel2019metalearners}, where the counterfactual outcomes are estimated using a plug-in estimator; and doubly robust (DR) method \citep{robins1994estimation, dudik2011doubly, zhou2018offline}, which combines the two approaches to obtain an unbiased estimator if either the counterfactual outcomes estimator or propensity score estimator is unbiased.
\citet{dudik2014doubly} shows that DR estimator is often more efficient in both policy evaluation and optimization tasks when propensity scores or the underlying demand is estimated reasonably well. We explore how these different methods relate to the class of loss functions we propose in \Cref{sec:ipw_relationship}. There is a recent stream of literature that applies some ideas from off-policy learning to pricing problems with censored demand, which occurs due to a lack of inventory \citep{ban2020confidence, bu2022offline, qi2022offline}. Instead of focusing on aggregate demand, which is continuous, we focus on a  personalized pricing setting, where the observed outcomes associated with individuals are binary (i.e., purchased or not). 

There are multiple variants of the IPS method, including normalization via reweighting \citep{lunceford2004stratification, austin2015moving} and trimming of the weights to reduce the variance of the estimates \citep{elliott2008model,ionides2008truncated}. When the logging policy is not known, one can learn balancing weights jointly without the need for a plug-in estimation of propensity scores \citep{kallus2018balanced,sondhi2020balanced}. %
For off-policy optimization, it is desirable to penalize actions that induce high variance in the loss function, %
as proposed in \cite{swaminathan2015counterfactual,joachims2018deep}, and \cite{bertsimas2018optimization}. This can also be thought of as a distributionally robust approach using KL-divergence~\citep{faury2020distributionally, si2020distributionally}. %
 One can also minimize the variance of the loss through reweighting or retargeting these loss functions as a function of $X$, the covariates \citep{kallus2020efficient}. However, such a procedure, where one introduces bias in the objective, is not appropriate for policy evaluation, which is studied in this paper. 

In this paper, we bridge gaps between learning from noisy supervision literature (i.e., the corrupted labels)  in the machine learning community and contextual pricing. This framework was initially used to analyze binary classification tasks with noisy labels \citep{natarajan2013learning} -- the outcomes that we would ideally observe (the \textit{clean labels}) are probabilistically transformed, resulting in \textit{corrupted labels} that are actually observed. In \cite{natarajan2013learning}, the true binary label switches signs with a known probability. This literature shows how the classification loss functions appropriate for the  \textit{clean labels} can be transformed to loss functions for observed data, which produce the same loss in expectation. The framework has been extended to study semi-supervised learning \citep{van2017theory}, learning with partial labels \citep{cour2011learning}, and identifying treatment responders \citep{kallus2019classifying}. This last work is especially relevant to this work, in that it applies ideas from corrupted labels to causal inference with binary actions or treatments. %
We extend this idea to a contextual pricing setting with multiple treatments but binary outcomes (purchase or no purchase), and we identify minimum variance and robust loss functions.

\section{Problem Formulation}

\subsection{Contextual Pricing - Customer Valuations, Observational Data, and Assumptions}

For personalized pricing, each customer is offered a contextualized price $P$ for a given product. We use $P \in \mathcal{P} $ to denote the price offered, which depends on the features of a customer and/or a product, $X \in \mathcal{X}$. After the price is prescribed, we observe the purchase decision $Y\in\{0,1\}$, where 1 corresponds to a purchase and 0 corresponds to no purchase. 
We assume prices are chosen from  a discrete price ladder $\mathcal{P} =\{p_1,\cdots,p_m\}$, where $m$ is the number of possible prices. This is a common assumption in the pricing literature \citep{cohen2017impact,biggs2021model}. %
The historical pricing policy $\pi_0(p|X)=\mathbb{P}(P=p|X)$ is assumed to be known, where $p$ represents a specific value to which $P$ is set. 
This assumption implies that the practitioners know or have recorded the pricing policies employed in the past. The assumption is likely to be satisfied when the firm is already implementing some form of algorithmic pricing. 
The customers' valuation or willingness to pay for that product is denoted as $V$, which is a heterogeneous distribution depending on $X$. For simplicity, we suppress the dependency of $X$ in $Y$ and $V$. 
With a discrete price ladder, the (potentially continuous) distribution of customer valuations can be equivalently modeled as a discrete distribution over $\mathcal{P} \cup \{ p_0 \}$, with a (possibly empty) set of customers with valuation $p_0$ who will not purchase at any price offered with $p_0 < p_1< ... <p_m$.

We observe whether a customer purchased the product at a given price, %
but do not observe the valuation of each customer. 
We refer to such data with $(X,P,Y)$ tuple as \textit{observational data}. 
Using the potential outcome framework \citep{rubin2005causal}, for a given price $p$, we observe a historical purchase decision $Y = Y(p)$ if $P=p$, and we cannot observe $Y(p')$ if $P\neq p'$. That is, we can only observe the purchase outcome for the price which was offered, but do not know what the purchase outcome would have been for other prices.
Here we make a simple observation that a customer will purchase if their valuation $V$ is greater than or equal to the price they are offered.

\begin{definition} \textit{Purchase outcomes}
\label{y_def}
\begin{equation}
Y(P) =    \begin{cases}
      1   & \text{if} ~~ P \leq V \\
      0   & \text{if} ~~  P > V    \\
    \end{cases} 
\end{equation}
\end{definition}

Accordingly, given $n$ observations, the observed data can be described as $\mathcal S_n = \{(X_i,P_i,Y_i)\}_{i=1}^n$, where $Y_i=Y_i(P_i)$ is the observed outcome. We note that Definition \ref{y_def} implies monotonicity in the price response, $Y(p) \geq Y(p'), ~ \forall p \leq p' ~\in \mathcal{P}$. This implies that customers are rational, that is, customers would not have purchased an item at the higher price if they chose not to purchase at a lower price.

For identifiability of conditional average treatment effect, we also require the following assumptions which are standard in the causal inference literature \citep{angrist1996identification,hirano2003efficient,pearl2010brief,swaminathan2015counterfactual,gao2021enhancing,gao2021human,gao2023learning}.
\begin{assumption}(Ignorability)
\label{ass:ig}
$Y(p) \perp   P | X, \quad \forall p ~\in \mathcal{P}$
\end{assumption}

\begin{assumption}(Overlap)
\label{ass:ovp}
$ \pi_0(p|X) > 0, \quad\forall p\in \mathcal{P}$
\end{assumption}

The ignorability assumption %
means that the pricing policy was chosen as a function of the observed covariates $X$ and that there are no unmeasured confounding variables that affect both the pricing decision and the purchase outcome. 
As pointed out in \cite{bertsimas2020predictive}, this assumption is particularly defensible in prescriptive analytics. Suppose actions represent historical managerial or algorithmic decisions that have been made based on observable quantities available to the decision-maker. As long as these quantities were also recorded as part of the feature, this assumption is guaranteed to hold. %

The overlap assumption requires that all prices have a nonzero chance of being offered to all customers historically, which implies the historic policy must be randomized to some extent. The known impossibility result of counterfactual evaluation applies when it is not satisfied~\citep{langford2008exploration}, and estimation of demand is possible only if further structural assumptions are made. 
If the overlap assumption is violated, practitioners usually make strong parametric assumptions such as linear demand or rely on the extrapolation power of nonparametric methods, which is arguably erroneous and unreliable compared to data with randomized assignment \citep{gordon2022close}. 
The overlap assumption is reasonable in many pricing settings where firms have adopted strategies that incorporate exploration. For example, the epsilon-greedy algorithm \citep{sutton2018reinforcement} can be built upon any deterministic pricing policy which  satisfies the overlap assumption.

For simplicity, we do not consider the effects of limited inventory or a finite selling horizon. We also study the pricing of a single item, in that we do not consider interaction effects from other items~\citep{gallego2014multiproduct,chen2019joint,chen2021nonparametric}. These are interesting avenues for future work but would result in significantly more complex policies.

\subsection{Loss Estimation for Pricing with Valuation Data}

In the setting where we observe each customer's valuation, it is straightforward to formulate a loss function that models the revenue that would have been obtained had a particular price been prescribed to a customer. More specifically, if the customer has a valuation that is greater than the price prescribed, then the customer will purchase and the revenue obtained will be the price prescribed. If the price we prescribe is greater than a customer's valuation, no purchase is made and the revenue associated with this price for the customer is zero. 
This rationale is captured in the following loss function with a slight modification of the loss function used in \cite{mohri2014learning}:

\begin{definition} 
\label{loss_pricing_random} Pricing loss function: $l_V(P,V) =   - P \mathbbm{1} \{ P \leq V \}.$
\end{definition}

\begin{figure}[!t]
	\centering
\begin{tikzpicture}
\begin{axis}[
    clip=false, 
    xmin=0,xmax=30,
    ymin=-20, ymax=5,
    grid=both,
    grid style={line width=.1pt, draw=gray!10},
    major grid style={line width=.2pt,draw=gray!50},
    axis lines=middle,
    minor tick num=5,
    enlargelimits={abs=0.5},
    axis line style={latex-latex},
    ticklabel style={font=\tiny,fill=white},
    xlabel style={at={(ticklabel* cs:1)},anchor=north west},
    ylabel style={at={(ticklabel* cs:1)},anchor=south west}
]
\coordinate (O) at (0.,0.);
\node[fill=white,circle,inner sep=0pt] (O-label) at ($(O)+(-135:-10pt)$) {$O$};
\draw[line width=0.7mm, orange, dotted] (0,0)--(15,-15);
\draw[line width=0.7mm, orange, dotted] (15,0)--(15,-15);
\draw[line width=0.7mm, orange, dotted] (15,0)--(30,0);
\draw (-2,5) node[left] {Loss $L_V(P,V)$};
\draw (30,-2) node[below] {Price offered $P$};
\draw[color={rgb:red,6;green,2;yellow,2}] (25,-10) node[below] {Valuation $V=15$};
\addplot [color=orange,only marks,mark size=2pt] table {
0 0
5 -5
10 -10
15 -15
20 0
25 0
30 0
};
\end{axis}
\end{tikzpicture}
\caption{Loss function when the valuation of a customer is known (i.e., $V=15$). X-axis and Y-axis represent the price offered and the loss (negative revenue) respectively. For different prices offered, we plot the corresponding loss with the orange dotted line.}
\label{fig:loss_function}
\end{figure}
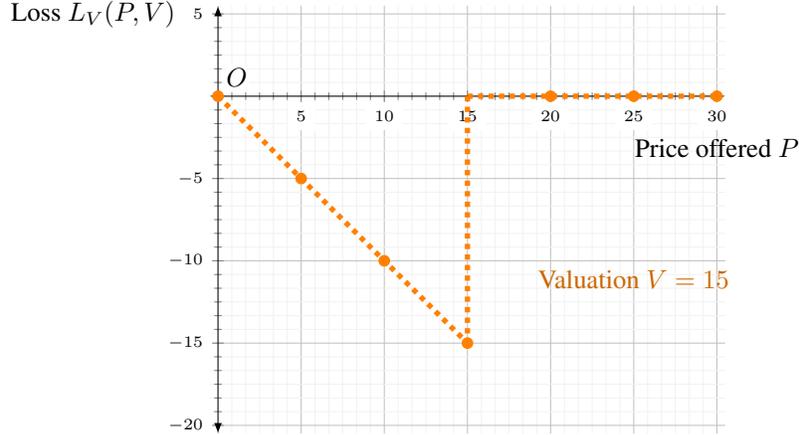
In keeping with machine learning conventions that a loss function is to be minimized, we define this as negative revenue. This is shown in Figure \ref{fig:loss_function}. An interesting observation about this loss function, in contrast to many other traditional loss functions (e.g., square loss, L1 loss), is its asymmetry, implying that it is better to underprice than to overprice. 

There are two broad tasks that a practitioner is typically interested in: i) accurate estimation of the loss function for a given policy,  and ii)  optimization of the loss function to obtain a new pricing policy. 
The former allows the evaluation of different pricing policies using historical data. %
Generally, being able to more accurately evaluate policies often also leads to better downstream decisions. %

We focus on evaluating a class of  contextual pricing policies $\pi \in \Pi$, a mapping $\mathcal{X} \rightarrow \Delta^{m}$, where $\pi(p|X)$ is the probability of assigning price $p$ to a customer with features $X$, and $\Delta^{m}=\{ \bm{f} \in [0,1]^m | \sum_{i=1}^m f_i = 1 \}$ is a probability simplex where the subscript $i$ indicates the $i$-th element of the vector. We note that deterministic pricing is a special case of this class. 
When evaluating a stochastic pricing policy, we can adapt the loss function from \Cref{loss_pricing_random} by taking expectation over the price offered:
\begin{equation}
\label{def:lv_v}
l_V(\pi,X,V) =    - \sum_{j=1}^m \pi(p_j|X)  p_j \mathbbm{1} \{ p_j \leq V \}. 
\end{equation}

This represents the expected negative revenue under the pricing policy $\pi$, given the customer valuation $V$ and features $X$. 
Using the principle of empirical risk minimization, the optimal pricing policy for minimizing the loss can be written as follows: %
\begin{equation}
\pi^* =\argmin_{\pi \in \Pi} \frac{1}{n} \sum_{i=1}^{n} l_V(\pi,X_i,V_i).
\end{equation}

\subsection{Connection between Customers' Valuations and Observed Data}
\label{sec:valuaiton_to_observed}
Although we do not observe customers' valuations, we do observe their purchase decisions.
Here we offer an important note that the observed outcome distribution (purchase or not at corresponding prices) is determined by the historical pricing policy and valuation distribution. 
We draw a parallel by using the corrupted labels framework from noisy supervision \citep{natarajan2013learning,van2017theory}  to describe this relationship between observed purchase decisions and latent valuations. 
Under this framework, the customer valuations are the \textit{clean labels}, which we would ideally observe but do not, while the observed outcomes (purchase or not at corresponding prices) are the \textit{corrupted labels}, which are a known probabilistic transformation of the clean labels.  
We define $\tilde{Y} \in \{0,1\}^{2m}$ as a one-hot encoding of the observed outcomes $(P,Y(P))$, such that
\begin{definition} 
\label{y_corr_def}
$ \tilde{Y} = [\tilde{Y}^{(1)},\tilde{Y}^{(0)}]$ where $\forall j \in \{1,...,m\},$
\begin{equation}
\tilde{Y}_{j}^{(1)} =    \begin{cases}
      1   & \text{if} ~~ P=p_j ~\text{and}~ Y(p_j)=1,\\
      0   & \text{if} ~~ \text{otherwise}      \\
    \end{cases}, 
    \qquad \tilde{Y}_{j}^{(0)} =    \begin{cases}
      1   & \text{if} ~~ P=p_j ~\text{and}~ Y(p_j)=0, \\
      0   & \text{if} ~~ \text{otherwise}      \\
    \end{cases} 
    ~~~ 
\end{equation}
\end{definition}

Denote $\bm{f_V} \in \Delta^{m+1}$ as the (discrete) distribution of valuations and $\bm{f_{\tilde{Y}}} \in \Delta^{2m}$ as the (discrete) distribution of observed outcomes, ordered according to Definition \ref{y_corr_def}:
\begin{align}
\bm{f_V}= &[\mathbb{P}(V=p_0|X),~\mathbb{P}(V=p_1|X), ~\mathbb{P}(V=p_2|X), \cdots, ~\mathbb{P}(V=p_m|X) ]  \label{eqn:fv}\\
\bm{f_{\tilde{Y}}}= [&\mathbb{P}(P=p_1,Y(p_1)=1|X), ~\mathbb{P}(P=p_2,Y(p_2)=1|X), \cdots, ~\mathbb{P}(P=p_m,Y(p_m)=1|X), \nonumber \\  &~\mathbb{P}(P=p_1,Y(p_1)=0|X), ~\mathbb{P}(P=p_2,Y(p_2)=0|X), \cdots, ~\mathbb{P}(P=p_m,Y(p_m)=0|X)] \label{eqn:fytilde}
\end{align}

The relationship between these distributions can be captured by a matrix $\bm{T} \in \mathbb{R}^{2m \times (m+1)}$, which depends on the historic pricing policy $\pi_0(P|X)$ and maps $\bm{f_V}$ to $\bm{f_{\tilde{Y}}}$ with $\bm{f_{\tilde{Y}}} = \bm{T}\bm{f_V}$. 
 We suppress the dependence on $X$ in the notation for $\bm{f_V}, \bm{f_{\tilde{Y}}}$, and $\bm{T}$ to streamline notation. 
 Formally, $\bm{T}$ can be defined as

\begin{definition} 
$\bm{T}=\begin{bmatrix} \bm{T}^{(1)} \\ \bm{T}^{(0)}  \end{bmatrix}$ where $ ~ \forall  i \in \{1 \cdots, 2m\},~ j \in \{1, \cdots, m+1\}$,
\begin{equation}
\bm{T}_{ij}^{(1)} =    \begin{cases}
      \pi_0(p_i|X)   & \text{if} ~~ j > i  \\
      0 & \text{otherwise}
    \end{cases}, \qquad
    \bm{T}_{ij}^{(0)} =    \begin{cases}
      \pi_0(p_i|X)   & \text{if} ~~ j \leq i \\
      0 & \text{otherwise}
    \end{cases}
\end{equation}
\end{definition}

\begin{example}(\textit{$\bm{T}$ Matrix that transforms valuation distribution to  price and sale joint distribution})

\centering 
\resizebox{\hsize}{!}{
$ \bm{T} = 
\begin{blockarray}{ccccccc}
 & \gra V= p_0 &  \gra V= p_1 &  \gra V= p_2  &  \gra \cdots &  \gra V=  p_{m-1} &  \gra V=  p_m \bla \\
\begin{block}{c(cccccc)}
\textcolor{gray}{(P=p_1 , Y(p_1)=1)}& 0 & \pi_0(p_1|X) & \pi_0(p_1|X)   & \cdots &  \pi_0(p_1|X) & \pi_0(p_1|X) \\
\textcolor{gray}{(P=p_2 ,  Y(p_2)=1)} & 0 & 0 & \pi_0(p_2|X)  & \cdots &  \pi_0(p_2|X) & \pi_0(p_2|X)\\
  \vdots & \vdots & \vdots  & \vdots & \cdots & \vdots & \vdots  \\
\textcolor{gray}{(P=p_{m-1} ,  Y(p_{m-1})=1)} &  0 & 0  & 0 & \cdots &  \pi_0(p_{m-1}|X) & \pi_0(p_{m-1}|X) \\
\textcolor{gray}{(P=p_m ,  Y(p_m)=1)} & 0 & 0 &   0 & \cdots &  0& \pi_0(p_m|X)  \\
\textcolor{gray}{(P=p_1 ,  Y(p_1)=0)} & \pi_0(p_1|X)  & 0  & 0  & \cdots & 0 & 0 \\
\textcolor{gray}{(P=p_2 ,  Y(p_2)=0)}& \pi_0(p_2|X) & \pi_0(p_2|X)  & 0   & \cdots & 0 & 0\\
   \vdots & \vdots & \vdots  & \vdots & \cdots & \vdots & \vdots  \\
\textcolor{gray}{(P=p_{m-1} ,  Y(p_{m-1})= 0 )} &  \pi_0(p_{m-1}|X)    & \pi_0(p_{m-1}|X)  & \pi_0(p_{m-1}|X)  & \cdots &  0 & 0 \\
\textcolor{gray}{(P=p_m ,  Y(p_m)= 0 )} &  \pi_0(p_m|X)  &  \pi_0(p_m|X)  &   \pi_0(p_m|X)  & \cdots &  \pi_0(p_m|X) & 0  \\ 
\end{block}
\end{blockarray}
$
 }
\end{example}
 
\begin{repeatexample} (\textit{continued})
 Each column in $\bm{T}$ corresponds to a potential observed outcome distribution of a customer with a given valuation, where the outcome is whether a purchase takes place at a prescribed price. %
 For example, if a customer has a valuation of $V=p_2$, we will observe an outcome where the customer purchases if prescribed a price $p_1$ ($P=p_1,Y(p_1)=1$) with probability $\pi_0(p_1|X)$, the historical probability a customer is assigned a price $p_1$. Conversely, it is not possible to observe $(P=p_3,Y(p_3)=1)$ for this customer, because they will not purchase if their valuation $p_2$ is less than the price they are offered $p_3$. Therefore, for column $V=p_2$, we may observe any outcome among $(P=p_1,Y(p_1)=1),(P=p_2, Y(p_2)=1), (P=p_3, Y(p_3)=0), ..., (P=p_m,Y(p_m)=0)$, but there is probability $0$  for all other outcomes that are not consistent with Definition \ref{y_def}.
\end{repeatexample}

With the transition matrix $\bm{T}$, we can use the valuation distribution and the historical pricing policy to calculate the observed sales distribution.

\begin{lemma}
\label{transformation_of_dist}
$\bm{f_{\tilde{Y}}}= \bm{T} \bm{f_V}$%
\end{lemma}

The proof of this Lemma can be found in Appendix \ref{transformation_of_dist} and follows from the definitions introduced so far, with the application of ignorability (Assumption \ref{ass:ig}). 
While this describes the relationship between the distribution of valuations and observed outcomes, it is not yet clear how to evaluate a pricing policy using the observed data, which we will address next.%

\subsubsection{Loss Estimation with Observed Data}
\label{sec:loss_estimation_obs}

With observational data, we do not observe the valuation, but instead, observe only the binary outcome of whether the item sold at the price offered. In such a setting, we cannot use the loss function given in \Cref{loss_pricing_random}.
We apply the theory of learning from corrupted labels to the contextual pricing setting to derive a suitable loss function that utilizes only the observational data. 
The rationale is that there exists a probabilistic transformation $\bm{R}$ that transforms the observed price/outcome distribution to the valuation distribution, so the valuation loss function can be used. Recall that in \Cref{sec:valuaiton_to_observed}, we derived a matrix $\bm{T}$ that transforms the valuation distribution to the observed outcomes $\bm{f_{\tilde{Y}}}= \bm{T} \bm{f_V}$. It can be shown that there exists a matrix $\bm{R} \in \mathbb{R}^{(m+1)\times 2m}$, which is able to make the transformation in the opposite direction, such that $\bm{f_V}= \bm{R} \bm{f_{\tilde{Y}}}$. For this transformation to be valid, we require that $\bm{RT}=\bm{I}$, i.e., $\bm{R}$ is a left inverse of $\bm{T}$. 
In addition, we define the loss vector under each possible valuation. 
If we consider all possible discrete customer valuations $V \in \{p_0,p_1,p_2, ... ,p_m\}$ that correspond to the discrete price ladder, the loss for all possible outcomes can be represented using a column vector, $\bm{l_V}^{\pi}$, which is shown below:
\begin{equation}
\label{def:lv_vec}
{\bm{l_V}^{\pi}} = [ l_V(\pi,X,p_0), l_V(\pi,X,p_1), l_V(\pi,X,p_2),..., l_V(\pi,X,p_m)]^{'}.
\end{equation}

We can apply this transformation to the valuation loss vector by using the following loss function:

\begin{definition}{\textit{Loss function for pricing with observational data:} }
$$l_{\tilde{Y}}(\pi,X,\tilde{Y})=\tilde{Y}' \bm{R}'\bm{l_V}^{\pi}~~~ \text{where}~~~ ~\bm{RT}=\bm{I}.$$
\end{definition}

Here $\tilde{Y}$ is the one-hot encoding of observed outcomes (\Cref{y_corr_def}).
When the expectation is taken over the observed outcomes, this loss function is equal to the expectation of the valuation loss as the following Lemma shows:

\begin{lemma}
\label{expected_loss}
$\mathbb{E}_{\tilde{Y}}[l_{\tilde{Y}}(\pi,X,\tilde{Y})|X] = \mathbb{E}_V[l_V(\pi,X,V)|X]$ for all $\bm{R}$ such that  $\bm{RT}=\bm{I}$.
\end{lemma}

\proof{Proof of Lemma \ref{expected_loss}:} 
$$\mathbb{E}_{\tilde{Y}}[\tilde{Y}'\bm{R}'\bm{l_V}^{\pi}|X]=\bm{f_{\tilde{Y}}}' \bm{R}' \bm{l_V}^{\pi}=\bm{f_V}' \bm{T}' \bm{R}'\bm{l_V}^{\pi}=\bm{f_V}'\bm{l_V}^{\pi}= \mathbb{E}_V[l_V(\pi,X,V)|X]$$
where the second equality follows from $\bm{f_{\tilde{Y}}}= \bm{T} \bm{f_V}$, while the third equality follows from $\bm{RT}=\bm{I}$.\Halmos
\endproof

This Lemma validates using $l_{\tilde{Y}}(\pi,X,\tilde{Y})=\tilde{Y}'\bm{R}'\bm{l_V}^{\pi}$ as the loss function in the corrupted data setting, as the expectation is the same as with valuation data. However, since the left inverse $\bm{R}$ is generally not unique ($2m \geq m + 1$), there exists a class of unbiased loss functions that are equivalent to the valuation in expectation, depending on the choice of $\bm{R}$. In \Cref{sec:min_var} and \Cref{sec:rob_est}, we will consider variants of $\bm{R}$ that lead to different loss functions with specific properties. With this loss function, we can then evaluate  a future policy $\pi$ using $\frac{1}{n} \sum_{i=1}^n l_{\tilde{Y}}(\pi, X_i , \tilde{Y}_i)$ or perform empirical risk minimization by $\min_{\pi\in\Pi}\frac{1}{n} \sum_{i=1}^n l_{\tilde{Y}}(\pi, X_i , \tilde{Y}_i)$ to find optimal pricing policy for the sample. Lastly, we  note that Assumption~\ref{ass:ovp}, requiring overlap of the historical pricing policy, ensures that a suitable $\bm{R}$ exists.

To illustrate how this loss function works on observational data and to help parse notation, we provide a toy example of the corrupted loss functions  with only two prices. Example \ref{worked_loss_function} shows one possible choice of $\bm{R}$ with a simple closed-form expression, which we will study in more detail in Section \ref{sec:CIPS}. 

\begin{example}{\textit{Simple worked example for a feasible $\bm{R}$ with two price options $ p_1, p_2$}}
\label{worked_loss_function}
\resizebox{\textwidth}{!}{
$
\bm{T} = 
\begin{blockarray}{cccc}
 & \gra V= p_0 & \gra V= p_1 & \gra V= p_2 \\
\begin{block}{c(ccc)}
\gra (P=p_1 , Y(p_1)=1) & 0 & \pi_0(p_1|X) & \pi_0(p_1|X)   \\
\gra (P=p_2 ,  Y(p_2)=1) & 0 & 0 & \pi_0(p_2|X) \\
\gra (P=p_1 ,  Y(p_1)=0) & \pi_0(p_1|X)  & 0  & 0   \\
\gra (P=p_2 ,  Y(p_2)=0)& \pi_0(p_2|X) & \pi_0(p_2|X)  & 0  \\
\end{block}
\end{blockarray}, ~~~~
\bm{R} = 
\begin{blockarray}{(cccc)}
0 & 0 & 1/\pi_0(p_1|X) & 0   \\
0 & 0   &  - 1/\pi_0(p_1|X) &  1/\pi_0(p_2|X) \\
0 & 1/\pi_0(p_2|X)  & 0 & 0  \\
\end{blockarray} \\
$
}

It can be verified that $\bm{RT}=\bm{I}$, so this choice of $\bm{R}$ is feasible and will result in an unbiased loss function. From \Cref{def:lv_v} and \Cref{def:lv_vec}:

\[ \bm{l_V}^{\pi} = 
\begin{blockarray}{(c)}
l_{V}(\pi,X,p_0) \\ l_{V}(\pi,X,p_1) \\ l_{V}(\pi,X,p_2) \\
\end{blockarray} ~=~
\begin{blockarray}{(c)}
- \sum_{j=1}^2 \pi(p_j|X)  p_j \mathbbm{1} \{ p_j \leq p_0 \} \\
- \sum_{j=1}^2 \pi(p_j|X)  p_j \mathbbm{1} \{ p_j \leq p_1 \} \\
- \sum_{j=1}^2 \pi(p_j|X)  p_j \mathbbm{1} \{ p_j \leq p_2 \} \\
\end{blockarray} ~=~
\begin{blockarray}{(c)}
 0 \\
- \pi(p_1|X)  p_1 \\
- \pi(p_1|X)  p_1 -  \pi(p_2|X)  p_2  \\
\end{blockarray}
\]

It follows that the loss $\bm{R}'\bm{l_V}^{\pi}$ associated with each potential observed outcome can be written as follows:
\begin{align}
\begin{blockarray}{(c)}
l_{\tilde{Y}}(\pi, X,P=p_1,Y(p_1)=1) \\
l_{\tilde{Y}}(\pi, X,P=p_2,Y(p_2)=1) \\
l_{\tilde{Y}}(\pi, X,P=p_1,Y(p_1)=0) \\
l_{\tilde{Y}}(\pi, X,P=p_2,Y(p_2)=0) 
\end{blockarray} 
& = 
\begin{blockarray}{(ccc)}
0 & 0 & 0   \\
0 & 0  & 1/\pi_0(p_2|X)  \\
1/\pi_0(p_1|X) & - 1/\pi_0(p_1|X)  & 0   \\
0 & 1/\pi_0(p_2|X)  & 0
\end{blockarray} 
~
\begin{blockarray}{(c)}
 0 \\
- \pi(p_1|X)  p_1 \\
- \pi(p_1|X)  p_1 -  \pi(p_2|X)  p_2  \\
\end{blockarray}\nonumber \\
& ~=~
\begin{blockarray}{(c)}
0 \\
-(\pi(p_1|X) p_1 + \pi(p_2|X) p_2)/ \pi_0(p_2|X)    \\
\pi(p_1|X) p_1   / \pi_0(p_1|X) \\
-\pi(p_1|X) p_1   / \pi_0(p_2|X) 
\end{blockarray} 
\end{align} 

We now show how to apply this loss function to a toy dataset. Suppose we have two observations in our sample: one customer with features $X_1$ purchased at a price $p_2$, while another customer with features $X_2$ did not purchase at a price $p_1$. These are respectively one-hot encoded according to \Cref{y_corr_def} as ${\tilde{Y}_1}^{'}=[0,1,0,0],~{\tilde{Y}_2}^{'}=[0,0,1,0]$. We can evaluate the pricing policy $\pi$ on the limited sample according to

\[ \frac{1}{n} \sum_{i=1}^n l_{\tilde{Y}}(\pi, X_i , \tilde{Y}_i) = \frac{1}{2} \sum_{i=1}^2 \tilde{Y}_i' \bm{R}'\bm{l_V}^{\pi} = \frac{1}{2}\left( -\frac{\pi(p_1|X_1) p_1 + \pi(p_2|X_1) p_2} {\pi_0(p_2|X_1) }  + \frac{\pi(p_1|X_2) p_1} { \pi_0(p_1|X_2) }\right) \]

We could also optimize to find the best pricing policy by solving
\[\pi^* =\argmin_{\pi \in \Pi} \frac{1}{n} \sum_{i=1}^n l_{\tilde{Y}}(\pi, X_i , \tilde{Y}_i)=  \argmin_{\pi \in \Pi} \frac{1}{2}\left( -\frac{\pi(p_1|X_1) p_1 + \pi(p_2|X_1) p_2} {\pi_0(p_2|X_1) }  + \frac{\pi(p_1|X_2) p_1} { \pi_0(p_1|X_2) }\right) \]

\end{example}

\bla

\section{Minimum Variance Estimator}
\label{sec:min_var}

To estimate the loss, one desirable property is that the loss function has a low variance. Even though all loss functions with $\bm{RT}=\bm{I}$ are equivalent in expectation (unbiased), different $\bm{R}$ matrices correspond to loss functions with different variances.\footnote{The loss functions from \Cref{expected_loss} are unbiased estimates of the valuation loss for a given policy, $\pi$. This should not be confused with whether optimizing the loss function will result in the optimal pricing policy, which requires that the policy class $\Pi$ contain the optimal policy. It may also be desirable to bias the policy to reduce variance; for example, see \cite{swaminathan2015counterfactual}. 
Such methods could  also be incorporated into the proposed approach and may be improved if a corresponding $\bm{R}$ is chosen carefully.} Choosing an unbiased loss function with low variance is desirable since it improves the  efficiency of the estimator and makes it less likely that a suboptimal policy will be chosen due to noise. An expression for the variance of the loss function, showing dependence on $\bm{R}$, is stated in the following result:

\begin{proposition}
$\Var[\tilde{Y}'\bm{R}'\bm{l_V}^{\pi}] = \mathbb{E}_X [\Var_{\tilde{Y}}(\tilde{Y}'\bm{R}'\bm{l_V}^{\pi}|X)] + \Var_X(\mathbb{E}_{\tilde{Y}}[\tilde{Y}'\bm{R}'\bm{l_V}^{\pi}|X])$\label{prop_total_variance}
\end{proposition}

This follows directly from the law of total variance. 
\Cref{prop_total_variance} shows that  the choice of $\bm{R}$ does not affect the second term, since $\mathbb{E}_{\tilde{Y}}[\tilde{Y}'\bm{R}'\bm{l_V}^{\pi}|X]$ is constant in $\bm{R}$, due to the unbiasedness of the estimator (\Cref{expected_loss}). Therefore, to choose the most efficient $\bm{R}$, we focus on the conditional variance in the first term. The conditional variance can be expressed as follows:

\begin{proposition}
\label{cov_prop}
$\Var_{\tilde{Y}}[\tilde{Y}'\bm{R}'\bm{l_V}^{\pi}|X]= \bm{l_V}^{\pi'}\bm{R}\bm{\Sigma}_{\tilde{Y}} \bm{R}'\bm{l_V}^{\pi}$ where 
\begin{equation}
	\bm{\Sigma}_{\tilde{Y}} = \Cov[\tilde{Y}|X]= \mathrm{diag}(\bm{f_{\tilde{Y}}})- \bm{f_{\tilde{Y}}}\bm{f_{\tilde{Y}}}',\label{cov_mat} \\ 
\end{equation} 
where $\mathrm{diag}(\bm{x})$ is a square matrix of zeros but with $x_i$ in position $(i,i)$ on the diagonal.
\end{proposition}

As a result, one can formulate the problem of finding the $\bm{R}$ matrix with the minimum asymptotic variance  as a constrained quadratic optimization problem. We prove that this optimization has a simple closed-form solution, as given in \Cref{min_var_soln}. 

\begin{lemma}
\label{min_var_soln}
The $\bm{R}$ matrix corresponding to the minimum variance can be determined by solving the following optimization problem: 
 \begin{align}
 \label{eqn:mvprob}
 \bm{R}_{MV}=  \argmin_R~& \Var_{\tilde{Y}}[\tilde{Y}'\bm{R}'\bm{l_V}^{\pi}|X]\\
 \text{s.t.}~ &\bm{RT}=\bm{I} \nonumber
 \end{align}
 The corresponding solution to \Cref{eqn:mvprob} is: 
 \begin{equation}
  \bm{R}_{MV}= (\bm{T}'\mathrm{diag} (\bm{f_{\tilde{Y}}})^{-1}\bm{T})^{-1}\bm{T}' \mathrm{diag}(\bm{f_{\tilde{Y}}})^{-1}  \label{MV_estimate}
 \end{equation}
\end{lemma}

The proof can be found in Appendix \ref{min_var_proof} and follows from solving the KKT conditions for the corresponding optimization problem. As $\bm{R}_{MV}$ depends on $X$, it must be calculated for individual customers. Lemma \ref{min_var_soln} is particularly useful, as the computational efficiency of calculating a closed-form solution rather than solving a complex optimization problem makes the approach tractable for large datasets. 

While this is the minimum variance loss function, it uses %
the distribution $\bm{f_{\tilde{Y}}}$, which is often not known a priori. However, if the practitioner has prior knowledge or intuition about the distribution $\bm{\hat{f}}_{\tilde{Y}}$ (e.g., at what prices the item usually does or does not sell), then they can use this prior to generate lower variance loss functions. We cover two scenarios in \Cref{sec:ipw_relationship}: one when an item is unlikely to be purchased at any price, and one when it is highly likely to be purchased. Alternatively, in many cases, it is possible to estimate $\bm{\hat{f}}_{\tilde{Y}}$ directly from data. In this case, the matrix can be calculated using a plug-in estimator:

$$\hat{\bm{R}}_{MV}=(\bm{T}'\mathrm{diag}(\bm{\hat{f}}_{\tilde{Y}})^{-1}\bm{T})^{-1}\bm{T}'\mathrm{diag}(\bm{\hat{f}_{\tilde{Y}}})^{-1}$$

In particular, $\bm{\hat{f}}_{\tilde{Y}}$  can be estimated using a classification algorithm with a proper loss function to estimate $\mathbb{P}(Y=1|X,P)$ and applying Bayes Theorem.\footnote{More specifically, given an observational dataset with $N$ samples $\{X,P,Y\}_{i=1}^N$, we can directly fit a machine learning model such as Logistic Regression or Neural Network, parameterized by $\theta$, and get $\mathbb{P}_\theta(Y=1|X,P)$, then $\mathbb{P}_\theta(Y=1,P=p|X) = \mathbb{P}_\theta(Y=1|X,P=p)\pi_0(p|X)$.} It is easy to verify that $\hat{\bm{R}}_{MV}$ is still unbiased, even if the estimate for $\bm{\hat{f}}_{\tilde{Y}}$ is incorrect. However, the efficiency of the loss function is tied to the accuracy of the plug-in estimator and can have high variance if the estimate for $\bm{\hat{f}}_{\tilde{Y}}$ is inaccurate. 
This motivates the need to find low-variance loss functions that perform well even if $\bm{f_{\tilde{Y}}}$ cannot be estimated accurately. %

\section{Locally Robust Estimator}
\label{sec:rob_est}

In many settings, it is difficult to accurately estimate the plug-in estimator $\bm{\hat{f}_{\tilde{Y}}}$ because the demand function estimation is challenging, and often such estimates are biased. 
As such, we intend to find a loss function that has the minimum asymptotic variance even if $\bm{f_{\tilde{Y}}}$ is unable to be estimated accurately.  Ideally, such a loss function will have good evaluation performance even for distributions that differ from the estimate $\bm{\hat{f}_{\tilde{Y}}}$, and in particular one chosen by an adversary from within an uncertainty set:

\begin{definition}{\textit{Locally Robust formulation:}}
\begin{subequations}
 \begin{align}
 \label{robust_optimization_problem_start}
 \min_R ~\max_{f_{\tilde{Y}}} ~ \Var_{\tilde{Y}}[\tilde{Y}'\bm{R}'\bm{l_V}^{\pi}|X] = \min_R ~\max_{f_{\tilde{Y}}} & ~ \bm{l_V}^{\pi'}\bm{R} (\mathrm{diag}(\bm{f_{\tilde{Y}}})- \bm{f_{\tilde{Y}}}\bm{f_{\tilde{Y}}}') \bm{R}'\bm{l_V}^{\pi}\\ 
 \text{s.t.}~ \bm{b} \leq \bm{f_{\tilde{Y}}^{(1)}} & \leq \bm{u}  \\ 
    \bm{f_{\tilde{Y}}^{(1)}} + \bm{f_{\tilde{Y}}^{(0)}} & =\bm{\pi_0} \\ 
 \bm{RT}&=\bm{I}  \label{robust_optimization_problem_end}
 \end{align}
 \end{subequations}
 \end{definition}

 \noindent where $\bm{\pi_0}= [\pi_0(p_1|X),\pi_0(p_2|X),...,\pi_0(p_m|X)]$. 
The expression for the variance is from \Cref{cov_prop}. In the inner maximization problem, the adversary chooses $\bm{f_{\tilde{Y}}}=[\bm{f_{\tilde{Y}}^{(1)}}, \bm{f_{\tilde{Y}}^{(0)}} ]$ to maximize the variance of the loss function in response to the user's choice of $\bm{R}$. Additionally, $\bm{f_{\tilde{Y}}}$ is constrained to be part of an uncertainty set, with upper bound $u_i$ and lower bounds $b_i$ on the probability we observe a sale at each price $p_i$ in our historical data. 
For consistency of the estimate, we require $\bm{f_{\tilde{Y}}^{(1)}} + \bm{f_{\tilde{Y}}^{(0)}} =\bm{\pi_0}$, which means that the probability of observing a sale or not with a given price has to sum to the probability that price was offered in the historic data. Similarly, we require $0 \leq b_i \leq  u_i \leq \pi_0(p_i|X)$. In the outer minimization problem, the user chooses $R$, which is robust to all distributions in this uncertainty set.

There are multiple ways to find suitable upper $\bm{u}$ and lower bounds $\bm{b}$ for the uncertainty set:
\begin{enumerate}[label=(\roman*)]
\item A data-driven approach constructs a confidence interval around $\bm{\hat{f}_{\tilde{Y}}}$ based on the uncertainty in the plug-in estimator, with the conservatism of the uncertainty set determined by the confidence level chosen (i.e., $\alpha = 90\%$ or $95\%$). This can be achieved for general nonparametric estimators through bootstrapping, whereby different estimators are trained from re-sampled datasets and the resulting uncertainty is captured in a confidence interval \citep{efron1994introduction}. Other approaches for finding confidence intervals for nonparametric estimators include Bayesian approaches \citep{gal2016dropout} and conformal prediction \citep{papadopoulos2002inductive,wang2022probabilistic}. 
These approaches have an advantage in that the uncertainty of the estimator is reflected in the width of the interval. When the plug-in estimate is accurate and confident, we will have narrow intervals, while more uncertain estimation will have larger intervals.

\item An alternative approach is to find the uncertainty set through cross-validation, based on model performance. For example, if we set the upper and lower bounds to be a fixed distance $c$ from the plug-in estimate $b_i= \max\{\hat{f}_{\tilde{Y}i} -c,0\}, u_i= \min\{\hat{f}_{\tilde{Y}i} + c,\pi_0(p_i|X)\}$, then $c$ can be chosen through cross-validation using empirical data. In our experiments for loss function evaluation, $c$ with the lowest empirical total variance is chosen, while $c$ with the minimum estimated loss is preferred for loss function optimization. While not necessarily capturing heteroskedasticity, this approach has the advantage that the confidence parameter $\alpha$ does not need to be set in advance.
\end{enumerate}

Formulation (\ref{robust_optimization_problem_start}-\ref{robust_optimization_problem_end}) is a robust quadratic program with polyhedral constraints, where both the decision variable $\bm{R}$ and adversarial variable $\bm{f_{\tilde{Y}}}$ are quadratic. In general, this class of problems can be reformulated as convex optimization problems \citep{ben2015deriving}; however, we can show that this particular formulation happens to have a simple closed-form solution:

\begin{theorem} {\textit{Locally Robust solution:}} $\forall~ i \in \{1,...,m\}$ define:
\label{robust_soln}
\begin{equation} 
f_{\tilde{Y}i}^{rob(1)} =    \begin{cases}
      b_i   & \text{if} ~~  b_i \geq \frac{1}{2} \pi_0(p_i|X)\\
      \frac{1}{2} \pi_0(p_i|X)   & \text{if} ~~ b_i \leq \frac{1}{2} \pi_0(p_i|X)\leq u_i     \\
      u_i   & \text{if} ~~  u_i \leq \frac{1}{2} \pi_0(p_i|X)\\
    \end{cases}, ~~
    f_{\tilde{Y}i}^{rob(0)} =    \begin{cases}
      \pi_0(p_i|X) -  b_i   & \text{if} ~~  b_i \geq \frac{1}{2} \pi_0(p_i|X)\\
      \frac{1}{2} \pi_0(p_i|X)   & \text{if} ~~ b_i \leq \frac{1}{2} \pi_0(p_i|X)\leq u_i     \\
      \pi_0(p_i|X) -  u_i   & \text{if} ~~  u_i \leq \frac{1}{2} \pi_0(p_i|X)\\
    \end{cases} \label{eq:local_robust}
\end{equation}

The solution to (\ref{robust_optimization_problem_start}-\ref{robust_optimization_problem_end}) is  $\bm{R}^{rob}=   (\bm{T}'\mathrm{diag}(\bm{f_{\tilde{Y}}}^{rob})^{-1}\bm{T})^{-1}\bm{T}' \mathrm{diag}(\bm{f_{\tilde{Y}}}^{rob})^{-1} $
 
\end{theorem}

The proof can be found in Appendix \ref{robust_proof}. The main idea behind the proof is that there exists a pair of solutions, $\bm{f_{\tilde{Y}}}^{rob}$ and $\bm{R}^{rob}$, such that $\bm{f_{\tilde{Y}}}^{rob}$ is the worst-case distribution given $\bm{R}^{rob}$, and $\bm{R}^{rob}$ is the minimum variance loss function given the distribution $\bm{f_{\tilde{Y}}}^{rob}$. We verify that each solution satisfies the respective KKT conditions for each optimization problem. %

To aid with understanding this solution, we first investigate a special case where the distribution $\bm{f}_{\tilde{Y}}$ is not restricted by the uncertainty set, other than requiring it to be a valid distribution, i.e., $\bm{b}=\bm{0}$ and $\bm{u}=\bm{\pi}_0$. This corresponds to the extreme case where the uncertainties about customers' purchase decisions are as high as possible, and it is generally useful when the practitioner is not at all confident in the plug-in estimate. We will call this the robust solution.

\begin{corollary} {\textit{Robust solution:}}
\label{cor:global_robust_soln} When $\bm{b}=\bm{0}$ and $\bm{u}=\bm{\pi}_0$,
$$\bm{R}^{rob}=   (\bm{T}'\mathrm{diag}(\bm{f_{\tilde{Y}}}^{rob})^{-1}\bm{T})^{-1}\bm{T}' \mathrm{diag}(\bm{f_{\tilde{Y}}}^{rob})^{-1} $$ where $\bm{f_{\tilde{Y}}}^{rob(1)}= \frac{1}{2}\bm{\pi_0}, \bm{f_{\tilde{Y}}}^{rob(0)}= \frac{1}{2}\bm{\pi_0}$ or equivalently, $\bm{f_V}=\frac{1}{2}\bm{e_1}+\frac{1}{2}\bm{e_{m+1}}$.
\end{corollary}

In this setting, the worst-case distribution active at optimality is that half of the customers would purchase at any price and half would not. Equivalently, half the customers have the highest valuation while half have the lowest. Intuitively, it makes sense that this is a high-variance scenario where an accurate estimation of the revenue is difficult. %

Similarly, the  worst-case distribution for the locally robust solution (Equation (\ref{eq:local_robust})) biases $\bm{f_{\tilde{Y}}}$ toward $\frac{1}{2}\bm{\pi_0}$ as much as the uncertainty set will allow.  If the lower bound $b_i$ is greater than $\frac{1}{2} \pi_0(p_i|X)$, then $f_{\tilde{Y}i}^{rob(1)}$ is set to $b_i$, which is as close as possible to $\frac{1}{2} \pi_0(p_i|X)$, while if the upper bound $u_i$ is less than $\frac{1}{2} \pi_0(p_i|X)$, then $f_{\tilde{Y}i}^{rob(1)}$ is set to $u_i$. If the confidence interval $(b_i,u_i)$ contains $\frac{1}{2} \pi_0(p_i|X)$, then this is what the distribution is set to. %
As with the minimum variance loss function, having a closed form is advantageous as it allows   exact and quick calculations of $\bm{R}^{rob}$ (which depends on $X$)  for individual customers.

\section{Relationship to Off-Policy Evaluation Algorithms }
\label{sec:ipw_relationship}

Another possible, but not yet widely adopted approach to contextual pricing is to use IPS methods \citep{rosenbaum1983central, rosenbaum1987model}, also known as off-policy bandit learning \citep{dudik2011doubly, swaminathan2015counterfactual, zhou2018offline}. We show that some popular methods from causal  inference literature, when applied to this setting, are special cases of the class of loss functions we propose. In particular, we can find $\bm{R}$ matrices such that $\tilde{Y}' \bm{R}'\bm{l_V}^{\pi}$ corresponds to the IPS estimator and DR estimator, gaining insights into when such methods work well in the contextual pricing setting. 

\subsection{Connection to Inverse Propensity Score Estimators}
\label{sec:IPS_results}
IPS methods weight each observed customer with a given action by the inverse of the probability that action was prescribed to that customer. This increases the weight given to unlikely actions, while decreasing the weight given to likely actions, creating a pseudo-randomized trial. %
This is formalized in the following definition:

\begin{definition}
\textit{Inverse propensity score estimator applied to the contextual pricing setting for data sample $(X, P, Y)$:}
\begin{equation}
\label{ips_estimator}
l_{\IPS}(\pi,X,\tilde{Y})=      
      \sum_{j=1}^m \dfrac{\pi(p_j|X)} {\pi_0(p_j|X)} p_j  Y   \mathbbm{1}\{P=p_j\},
\end{equation}
 and the empirical revenue estimation on $\{X_i,P_i,Y_i\}_{i=1}^n$ is $\frac{1}{n}\sum_i l_{\IPS}(\pi,X_i,\tilde{Y}_i)$. 
\end{definition}

We prove in Lemma \ref{IPS_lemma} that there exists an $\bm{R}_{\IPS} \in \mathbb{R}^{(m+1)\times 2m}$ matrix that is equivalent to the IPS estimator in the contextual pricing setting. We note that relative to other $\bm{R}$ matrices, this has a relatively simple structure that can be expressed in closed form:

\begin{definition} 
\label{IPS_matrix_definition}
$\bm{R}_{\IPS}$ Matrix: 
\begin{align}
   &\textit{ First row:} \forall j \in \{1, \cdots, m\}: & \textit{Remaining rows:} \forall  i  \in \{2, \cdots, m+1\} ,~ j \in \{1, \cdots, m\}: \nonumber \\
    &[\bm{R}_{\IPS}]_{1,j} =    \begin{cases}
          1- 1/ \pi_0(p_1|X)   & \text{if} ~ j=1 \\
       1  & \text{if} ~ j>1 
    \end{cases}   
    &[\bm{R}_{\IPS}]_{i,j} =    \begin{cases}
      -1/\pi_0(p_j|X)   & \text{if} ~ i = j\\
      1/\pi_0(p_j|X)   & \text{if} ~  i = j+1 \\
      0 & \text{otherwise} 
    \end{cases}
\end{align}
\end{definition}

\begin{example} \textit{To help parse this notation, here is an example of $\bm{R}_{\IPS}$ in matrix form:}
$$\bm{R}_{\IPS}= \begin{bmatrix} 
 1-1/ \pi_0(p_1|X) & 1 & 1   & \cdots &  1 & 1 & 1 & \cdots &  1\\
 1/ \pi_0(p_1|X) & -1/ \pi_0(p_2|X) & 0  & \cdots &  0 & 0& 0 & \cdots &  0\\
  0 & 1/ \pi_0(p_2|X) & -1/ \pi_0(p_3|X)  & \cdots &  0 & 0& 0 & \cdots &  0\\
  0 & 0 & 1/ \pi_0(p_3|X)  & \cdots &  0 & 0& 0 & \cdots &  0\\
  \vdots & \vdots & \vdots  & \ddots & \vdots & \vdots & \vdots & \ddots & \vdots \\
 0 & 0  & 0 & \cdots &  1/ \pi_0(p_{m-1}|X)  & -1/ \pi_0(p_m|X)& 0 & \cdots &  0\\
 0 & 0 &   0 & \cdots &  0& 1/ \pi_0(p_m|X)& 0& \cdots &  0 \\
\end{bmatrix}$$
\end{example}

\begin{lemma} Loss function with $\bm{R}_{\IPS}$ corresponds to the IPS estimator
\label{IPS_lemma}
$$ \tilde{Y}'\bm{R}_{\IPS}' \bm{l_V}^{\pi} = -l_{\IPS}(\pi, X, \tilde{Y} )$$
\end{lemma}

The proof can be found in Appendix \ref{IPS_lemma_proof}. We note that there is a negative sign to transform the IPS estimator to a loss function to be minimized. %
While this result is not surprising since IPS estimators are unbiased estimators when the propensity scores are known, it is useful as a step toward proving the next result, relating IPS to the contextual pricing setting. We show that the IPS estimator can be viewed as a minimum variance loss function (see Section \ref{sec:min_var}) for a particular distribution of customer valuations:

\begin{lemma} Worst-case valuation distribution for IPS:
\label{lemma_IPS_min_var}
 \begin{align}
 \bm{R}_{\IPS} = & \argmin \Var_{\tilde{Y}}[\tilde{Y}'R'\bm{l_V}^{\pi}|X]\\
 \text{s.t.}~ & \bm{R}_{\IPS} \bm{T} = \bm{I} \nonumber
 \end{align}
 
for $\bm{f_{\tilde{Y}}}=\bm{T}\bm{e_1}$, or equivalently, $\bm{f_V} = \bm{e_1}$. 
\end{lemma}

This Lemma can be verified by showing that $\bm{R}_{\IPS}$ satisfies the KKT conditions for this optimization problem and can be found in Appendix \ref{IPS_min_proof}. The distribution for which the IPS method is a minimum variance loss function is $\bm{f_V} = \bm{e_1}$, which corresponds to all customers having a valuation of the minimum price. In this scenario, no customers will purchase at any price offered. While this extreme scenario is unlikely to occur in reality, a useful implication for practitioners is that the IPS estimator is likely to work well when \textit{the purchase probabilities are low}. %
We verify this experimentally in \Cref{sec:ipsvarying}.

\subsection{An Alternative when Purchase Probabilities Are High}
\label{sec:CIPS}
 We propose an alternative loss function, which is similar to IPS but will produce low variance estimates when the observed purchase probabilities are high. We define this as the Complementary Inverse Propensity Scoring estimator (CIPS), $\bm{R}_{\CIPS}$.

\begin{definition} 
$\bm{R}_{\CIPS}$ Matrix: \\

\begin{minipage}{.5\linewidth}
\textit{Last row:} $ j \in \{1, \cdots, m\}:$
\begin{equation} 
[\bm{R}_{\CIPS}]_{m+1,j} =    \begin{cases}
      1  & \text{if} ~  j > 1 \\
      1 -\frac{1}{\pi_0(p_m|X)}   & \text{if} ~  j=2m  \\
      0 & \text{otherwise} \nonumber 
    \end{cases}
    \end{equation}
\end{minipage}
\begin{minipage}{.5\linewidth}
\textit{First m rows:} $\forall  i \in  \{1, \cdots, m\} ,~ j \in \{1, \cdots, 2m\}$:
\begin{equation}
[\bm{R}_{\CIPS}]_{i,j} =    \begin{cases}
      \frac{1}{\pi_0(p_{j-m}|X)}   & \text{if} ~ i  = j -m  \\
      \frac{-1}{\pi_0(p_{j-m}|X)}   & \text{if} ~  i  = j-m+1    \\
      0 & \text{otherwise} 
    \end{cases} 
\end{equation}
\end{minipage}
\end{definition}

\begin{example}  \textit{To help parse notation, here is an example of $\bm{R}_{\CIPS}$ in matrix form:}
$$\bm{R}_{\CIPS}= \begin{bmatrix} 
 0 & \cdots &  0 &1/ \pi_0(p_1|X) & 0 & 0   & \cdots &  0 & 0 \\
  0 & \cdots &  0 & -1/ \pi_0(p_1|X) & 1/ \pi_0(p_2|X) & 0  & \cdots &  0 & 0\\
  0 & \cdots &  0 &0 & -1/ \pi_0(p_2|X) & 1/ \pi_0(p_3|X)  & \cdots &  0 & 0\\
  0 & \cdots &  0& 0 & 0 & -1/ \pi_0(p_3|X)  & \cdots &  0 & 0\\
  \vdots & \ddots & \vdots & \vdots & \vdots & \vdots  & \ddots & \vdots & \vdots  \\
 0 & \cdots &  0 & 0 & 0  & 0 & \cdots & - 1/ \pi_0(p_{m-1}|X)  & 1/ \pi_0(p_m|X)\\
1& \cdots &  1 & 1 & 1 &   1 & \cdots &  1 & 1- 1/ \pi_0(p_m|X) \\
\end{bmatrix}$$
\end{example}

We will show that this reduces to a much simpler form that does not require any matrix notation.

\begin{definition}
\textit{CIPS estimator applied to the contextual pricing setting:}
\begin{equation}
    l_{\CIPS}(\pi,X,\tilde{Y})=      
     \sum_{j=1}^{m} p_j \pi(p_j|X) - \sum_{j=1}^{m} p_j \pi(p_j|X)\dfrac{(1-Y) \mathbbm{1}\{P=p_j\} } {\pi_0(p_j|X)} 
     \label{eqn:cips}
\end{equation}
and the empirical revenue estimation on $\{X_i,P_i,Y_i\}_{i=1}^n$ is $\frac{1}{n}\sum_i l_{\CIPS}(\pi,X_i,\tilde{Y}_i)$. 
\end{definition}

\begin{lemma}
\label{CIPS_lemma} Loss function with $\bm{R}_{\CIPS}$ corresponds to the CIPS estimator
$$ \tilde{Y}'\bm{R}_{\CIPS}' \bm{l_V}^{\pi} = - l_{\CIPS}(\pi, X, \tilde{Y} )$$
\end{lemma}

The CIPS estimator in \Cref{eqn:cips} can be intuitively explained as the ideal revenue under policy $\pi$ minus the \textit{missed} revenue in terms of lost sales. The ideal revenue, which is the first term in \Cref{eqn:cips}, represents the revenue we would get if the customer would purchase at any price under policy $\pi$. 
The second term estimates the expected revenue the customer chooses not to spend in the historical data when averaging over all $X$. The only difference between the IPS estimator and the second term in the CIPS estimator is $Y$ and $1-Y$, where the IPS estimator calculates the expected revenue, and the CIPS estimator computes the expected missed revenue. When combined with the first term (i.e., ideal revenue - missed revenue), the CIPS estimator also offers an unbiased estimation for the expected revenue under $\pi$ (proved as part of \Cref{CIPS_corollary}). 
While the IPS and CIPS estimators both offer unbiased estimations, they result in different estimators for a finite sample. 

The IPS estimator  uses only the data samples where historical customers made a purchase. This suggests that when the propensity scores are known (i.e., the historical pricing policy), then it is possible to form estimates even in settings with incomplete data, such as sales data, where only the purchase decisions are observed. 
If the population distribution of the covariates $X$ is known (e.g., online stores track all incoming customers' data via cookies), CIPS, on the other hand, only uses the part of data samples where customers did not make a purchase, which may be more efficient when most of the customers make a purchase. 
As a result, IPS and CIPS estimators have different finite-sample performance and asymptotic variance, which we shall see experimentally in \Cref{sec:experiments}. %
As with the IPS estimator, we can identify the distribution of customer valuations for which CIPS is a minimum variance estimator. 
\bla 

\begin{lemma} Worst-case valuation distribution for CIPS:
\label{CIPS_corollary}
 \begin{align}
 \bm{R}_{\CIPS} = & \argmin \Var_{\tilde{Y}}[\tilde{Y}'R'\bm{l_V}^{\pi}|X]\\
 \text{s.t.}~ & \bm{R}_{\CIPS}\bm{T}  = I \nonumber
 \end{align}
 for $\bm{f_{\tilde{Y}}}=\bm{T}\bm{e_{m+1}}$, or equivalently $\bm{f_V} = \bm{e_{m+1}}$. 
\end{lemma}

The proofs for \Cref{CIPS_lemma} and \Cref{CIPS_corollary} are similar to  \Cref{IPS_lemma} and \Cref{lemma_IPS_min_var}, and are included for completeness in Appendix \ref{CIPS_lemma_proof} and Appendix \ref{CIPS_min_var_proof}. CIPS is a minimum variance loss function when the valuation distribution is $\bm{f_V} = \bm{e_{m+1}}$, that is, all customers have the highest valuation. In this setting, all customers will purchase at any price they are offered. As a practical insight, this performs well when the selling probabilities are high -- a direct contrast to the IPS estimator. %

\subsection{Connection to Doubly Robust Estimators}

The DR estimator \citep{dudik2011doubly, zhou2018offline}  combines the IPS method with an estimate of the reward.
Note that for the IPS estimator in \Cref{ips_estimator}, %
there is only a contribution to the reward estimate for the action that was observed for a customer. For this action, the DR estimator uses an estimate similar to IPS, but for the remaining unobserved actions, a plug-in estimator $\hat{\mu}(X, p_j)$ is used to estimate the reward. In this setting, $\hat{\mu}(X,p_j)$ is the estimated reward from giving a customer with features $X$ price $p_j$, and it may be estimated using machine learning models. It can also be easily adapted from an estimate of $\bm{\hat{f}_{\tilde{Y}}}$.

\begin{definition}
\textit{Doubly robust estimator applied to the contextual pricing setting:}
\begin{equation}
l_{DR}(\pi, X, \tilde{Y} )=    
       \sum_{j=1}^m \left( \hat{\mu}(X, p_j) \pi(p_j|X) + \frac{\pi( p_j|X)}{\pi_0(p_j|X)} ( p_j Y - \hat{\mu}(X, p_j)) \mathbbm{1}\{P=p_j\}  \right) 
\end{equation}

\noindent where $\hat{\mu}(X, p_j)=  p_j \hat{\mathbb{P}}(Y(p_j)=1| X, P=p_j )$.
The empirical revenue estimation on $\{X_i,P_i,Y_i\}_{i=1}^n$ is $\frac{1}{n}\sum_i l_{DR}(\pi,X_i,\tilde{Y}_i)$. 
\end{definition}

Although it takes a very different form, we can show that in the following result, the DR estimator is equivalent to the loss function using the minimum variance matrix, $\bm{\hat{R}}_{MV}$.

\begin{theorem} 
\label{DR_theorem} Doubly robust equivalence to the minimum variance estimator
$$ \tilde{Y}' \bm{\hat{R}}_{MV}' \bm{l_V}^{\pi} = - l_{DR}(\pi, X, \tilde{Y} )$$ where $\hat{\bm{R}}_{MV}=(\bm{T}'\mathrm{diag}(\bm{\hat{f}}_{\tilde{Y}})^{-1}\bm{T})^{-1}\bm{T}'\mathrm{diag}(\bm{\hat{f}_{\tilde{Y}}})^{-1}$ %
\end{theorem}

Typically, the DR approach is used when the propensity scores are unknown and need to be estimated. Its use is justified by the argument that the DR estimator is unbiased when either the propensity scores or direct reward estimates are unbiased \citep{dudik2011doubly,zhou2018offline}, which we examine in more detail in Appendix \ref{unknown_pricing_policy}. 
We provide an alternative justification, by analyzing the variance of the estimate when the propensity scores are known. We show that out of the class of proposed unbiased estimators, the variance is minimized by the DR estimator in the contextual pricing setting with discrete actions. 
Here the binary outcomes are important for enabling the characterization of the variance as a matrix. It is known in the causal inference literature that the DR estimator has the minimum asymptotic variance with binary actions \citep{robins1994estimation,cao2009improving}, but to the best of our knowledge, there are no such results in the multi-action case with binary outcomes. 
This proof bridges a gap between counterfactual risk minimization and learning with noisy supervision/corrupted labels, i.e., deriving popular estimators from the off-policy learning literature using learning with corrupted labels techniques.

\subsection{Generalization Bounds}
\label{sec:generalization_bounds}

In practice,  a policy is evaluated  on a finite sample of data. It is of interest to know how close the finite sample estimate is to the expected value, and how this scales with the size of our sample. 
Generalization bounds are well studied in the off-policy learning literature, with results established for IPS methods \citep{swaminathan2015counterfactual, zhou2017residual} and DR approaches \citep{zhou2018offline, kallus2018balanced}. We show that the generalization bounds from \cite{swaminathan2015counterfactual} can easily be extended to our setting. The bounds in \cite{swaminathan2015counterfactual} are an application of the empirical Bernstein bounds from  \cite{maurer2009empirical}.

To achieve these bounds, we follow \cite{swaminathan2015counterfactual} and introduce minor modifications to our estimation procedure. \cite{swaminathan2015counterfactual} use clipped estimators, which are a common approach for IPS that limits the potentially high variance when the propensity score is very small \citep{faury2020distributionally, si2020distributionally}. A clipped estimator is defined as:

\begin{equation*}
l_{\tilde{Y}}^M(\pi, X_i, \tilde{Y}_i) = \min \{ M,\tilde{Y}_i'\bm{R}'\bm{l_{V_i}}^{\pi} \}, ~~
\hat{l}_{\tilde{Y}}^M(\pi) = \frac{1}{n}\sum_{i=1}^n \min \{ M,\tilde{Y}_i'\bm{R}'\bm{l_{V_i}}^{\pi}),
\end{equation*}

 \noindent where $M$ is a threshold to avoid large propensity weights. For the purposes of proving the generalization bounds, it is also useful to introduce an auxiliary class of scaled loss functions, $\mathcal{G}_{\Pi}^M = \{ g_{\pi}^M: \mathcal{X} \times \{0,1\}^{2m} \rightarrow [0,1]\}$, where each policy $\pi \in \Pi$ corresponds to a function:
\begin{equation*}
    g_{\pi}^M(X_i,\tilde{Y}_i)= 1+ \frac{\min \{ M,\tilde{Y}' \bm{R}'\bm{l_{V_i}}^{\pi}\}}{M} 
\end{equation*}

This class of loss functions has range $[0,1]$, which allows generalization results from classification literature to be applied \citep{maurer2009empirical}. As in \cite{swaminathan2015counterfactual}, we use the notion of the covering number to measure the complexity of this function class (see \cite{maurer2009empirical} for a detailed treatment). 

\begin{definition} (Covering numbers)
\begin{itemize}
\item $\epsilon$ cover $\mathcal{N}(\epsilon,A,||\cdot||_{\infty})$: the size of the smallest cardinality subset $A_0 \subseteq A \subseteq R^n$ such that A is contained in the union of balls of radius $\epsilon$ centered at points in $A_0$, in the metric induced by $||\cdot||_{\infty}$. 
\item Covering number $\mathcal{N}_{\infty}(\epsilon,\mathcal{G},||\cdot||_{\infty}) =  \underset{(X_i,\tilde{Y}_i) \in  (\mathcal{X} \times \{0,1\}^{2m} ) }{\sup}  \mathcal{N}(\epsilon, \mathcal{G}(\{(X_i,\tilde{Y}_i)\}),n)$ where $\mathcal{G}(\{(X_i,\tilde{Y}_i)\}) = \{g(X_1,\tilde{Y}_1),..., X_n,\tilde{Y}_n)) : g \in \mathcal{G} \}$ is the function class evaluated at a sample.
\end{itemize}
\end{definition}

The covering number can also be upper bounded by a function of the VC dimension (see \cite{wellner2013weak}), so that can also be used as an alternative. Using the covering number, and denoting the empirical variance of the loss function evaluated on the sample (i.e., $\frac{1}{n}\sum_{i=1}^n ( l_{\tilde{Y}}^M(\bm{\pi}(X_i), \tilde{Y}_i') - \hat{l}_{\tilde{Y}}^M(\pi))^2$) by $\Var_n(\hat{l}_{\tilde{Y}}^M(\pi))$,
we can derive generalization bounds for the loss function we propose:

\begin{theorem} \label{generalization_bound_IPS}
Define $\mathcal{M}(n)=10 \mathcal{N}_{\infty}(\frac{1}{n},\mathcal{G}_{\Pi}^M ,2n)$ for a stochastic hypothesis space $\Pi$. For $n\geq 16$, with probability at least $1-\gamma$, 
\begin{equation*}
    \forall \pi \in \Pi, ~~ \mathbb{E}_{V,X}[l_V(\pi,X,V)] - \hat{l}_{\tilde{Y}}^M(\pi) \leq  \sqrt{\frac{18 \Var_n(\hat{l}_{\tilde{Y}}^M(\pi)) \log(\mathcal{M}(n)/\gamma)}{n}} + \frac{15M  \log(\mathcal{M}(n)/\gamma) }{n-1}.
\end{equation*}
\end{theorem}

This bound shows that the sampled loss function converges at the rate $O(1/\sqrt{n})$. This bound also gives us additional insight into why our loss function works well relative to the IPS estimators. The variance term $\Var_n(\hat{l}_{\tilde{Y}}^M(\pi))$ will typically be smaller with a carefully chosen $R$ (see Lemma \ref{min_var_soln}), compared to $R_{IPS}$,  which has a near-optimal variance if all customers have a valuation equal to the lowest price, as shown in Lemma \ref{lemma_IPS_min_var}.\footnote{Although these lemmas apply to the asymptotic variance, we also show experimentally that the variance is reduced for finite samples.} We  also note the dependence of the bound on $\mathcal{N}_{\infty}(\epsilon,\mathcal{G},||\cdot||_{\infty})$. This suggests that the difference between the sample and expected value grows with the complexity of the function class, as is common in statistical learning. This is an example of a bias-variance trade-off, where a richer function class will have greater variance, but a lower bias.

Matching our setting, \cite{swaminathan2015counterfactual} assume that the historical logging policy/propensity scores are known. 
In the setting where these have to be estimated along with the estimated reward,  \cite{zhou2018offline} provide a similar $O(1/\sqrt{n})$ bound using a DR estimator. As the DR estimator is equivalent to the minimum variance estimator we propose, its generalization bounds would apply to that estimator.

\section{Experiments}
\label{sec:experiments}

We first compare the algorithms in terms of loss function evaluation accuracy. For this task, the objective is to get the most accurate estimate of the revenue for a future policy based only on existing observational data. To achieve this, we measure the mean square error (MSE) between the estimated loss and the true loss of a given policy. 
In addition, we evaluate the algorithms on loss function optimization, where the goal is to prescribe the policy that generates the highest expected revenue. For this task, we  report the final reward corresponding to the optimized policy.

\subsection{Synthetic Data Experiments}
\label{sec:syndata}
We perform extensive experiments using synthetic data, where the underlying distributions are known and counterfactual outcomes can be generated. In our synthetic experiments, feature $X$ is sampled from a uniform distribution $\mathcal{U}(-1,1)^{d}$, where $d$ is the feature dimension. We study a setting with customer heterogeneity where the purchase outcome is generated using  $\mathbb{P}(Y=1|X=\bm{x},P=p) = \sigma(|\bm{w_1' x}| - |\bm{w_2' x}| p)$, where each weight $w_{1i}, w_{2i},~ i \in [d],$ is sampled from a uniform distribution $\mathcal{U}[0,1]$, $\sigma(x) = \frac{1}{1+e^{-x}}$ is the sigmoid function, and $p$ is the price assigned. 
The absolute value transformation for the price coefficient is used to ensure demand monotonicity with respect to price (decreases while price increases, $p > 0$). 
The historical pricing policy (logging policy) is set as $\text{softmax}(\{\mathbb{P}(Y=1|X=\bm{x}, P=p_1),\mathbb{P}(Y=1|X=\bm{x}, P=p_2),\cdots,\mathbb{P}(Y=1|X=\bm{x}, P=p_m)\} * \lambda)$, where $\lambda$ %
is set as 5 and the softmax function is $\text{softmax}(\textbf{x})_j=\frac{e^{x_j}}{\sum_i e^{x_i}}$, and the subscript $j$ indicates the $j$-th element of the vector. This approach of setting the logging policy is common in the off-policy learning literature \citep{swaminathan2015counterfactual} and corresponds to a historical policy that can capture some demand signal but still has some stochasticity in which action is chosen. Prices are chosen from a price grid of  \{1,2,3,4,5\}. For policy optimization, all estimators are optimized using the Adam optimizer~\citep{kingma2014adam}.

\subsubsection{Loss Function Evaluation and Optimization with Varying Demand Accuracy}
\label{sec:varying_accuracy}

We begin by studying the setting where the accuracy of the estimated demand  is manually controlled to explore the performance of the proposed methods. The demand estimator is set as a linear interpolation between the true probability of sale $\mathbb{P}_\text{true}$ and $\mathbb{P}_\text{0}$, where $\mathbb{P}_\text{0}$ represents a poor estimate of the true purchase probabilities. %
More specifically, $\mathbb{P}_{0}$ states the purchase probabilities of all customers at every price point as 0.01, which  is intentionally set to be incorrrect %
for this dataset. To generate the demand function, we set $\mathbb{P}(Y=1|X=x, P=p) = \alpha * \mathbb{P}_\text{true}(Y=1|X=x, P=p) + (1-\alpha) * \mathbb{P}_\text{0}(Y=1|X=x,p=p)$. Here, $\alpha$ represents the demand estimation quality. When $\alpha=1$, we recover the true demand, and when $\alpha=0$, the demand estimates that everything sells with a probability of 0.01. Later we also present results with demand estimated using machine learning models in \Cref{sec:demand_estimated}.

To generate a  policy $\pi$ to be evaluated, we follow the approach from \cite{swaminathan2015counterfactual}: for each repetition, we train a multi-label logistic regression model (fit one model for each price) on a small dataset with size $N_{tr} = 100$ to estimate demand. To generate the policy, we choose the price with the highest estimated revenue on each data instance in the dataset. As a result, the policy we are evaluating can be considered to be generally aligned with high revenues, but with some stochasticity. For policy evaluation, the metric of interest is defined as the mean squared error $\frac{1}{N}\sum(\frac{1}{n}\sum_i l_V(\pi,X_i,V_i)- \frac{1}{n}\sum_i l_{\tilde{Y}}(\bm{\pi}(X_i),\tilde{Y}_i))^2$, where $N$ is the number of repetitions, $n$ is the size of the dataset, $l_{\tilde{Y}}(\bm{\pi}(X),\tilde{Y}_i)$ is the estimated loss function, and $l_V(\bm{\pi}(X),V_i)$ is known in the synthetic setting. 

We compare several benchmarks  in our experiments, including Inverse Propensity Score (IPS), Minimum Variance / Doubly Robust (DR), Robust and Locally Robust (LOCAL ROB) estimators. Following our discussion on selecting the uncertainty interval for the Locally Robust estimator in \Cref{sec:rob_est}, we select the uncertainty set by choosing a fixed $c$ using a validation set. The uncertainty set is $b_i= \max\{\hat{f}_{\tilde{Y}i} -c,0\}, u_i= \min\{\hat{f}_{\tilde{Y}i} + c, \pi_0(p_i|X)\}$. The selection is based on the total variance as evaluated using the validation set. This is justified by \Cref{prop_total_variance}, since the selection of $\bm{R}$ based on the conditional asymptotic variance is equivalent to selecting based on the total variance. 

\begin{figure}[!t]
	\centering
	\subfloat[n = 50]{\includegraphics[width=0.5\textwidth]{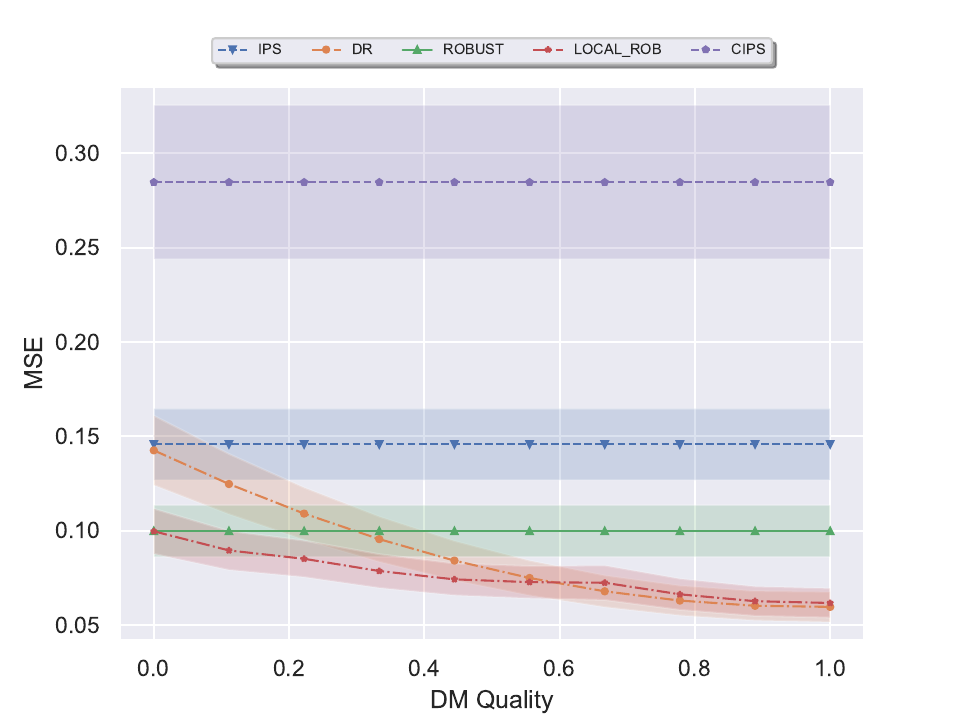}}
	\subfloat[n = 100]{\includegraphics[width=0.5\textwidth]{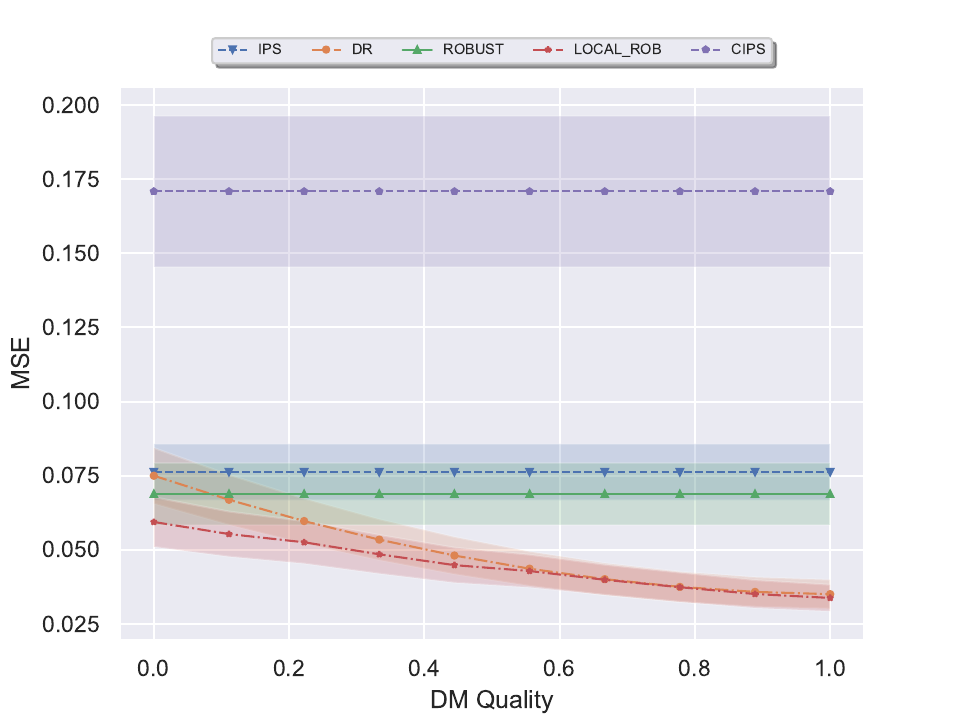}}\\
	\subfloat[n = 200]{\includegraphics[width=0.5\textwidth]{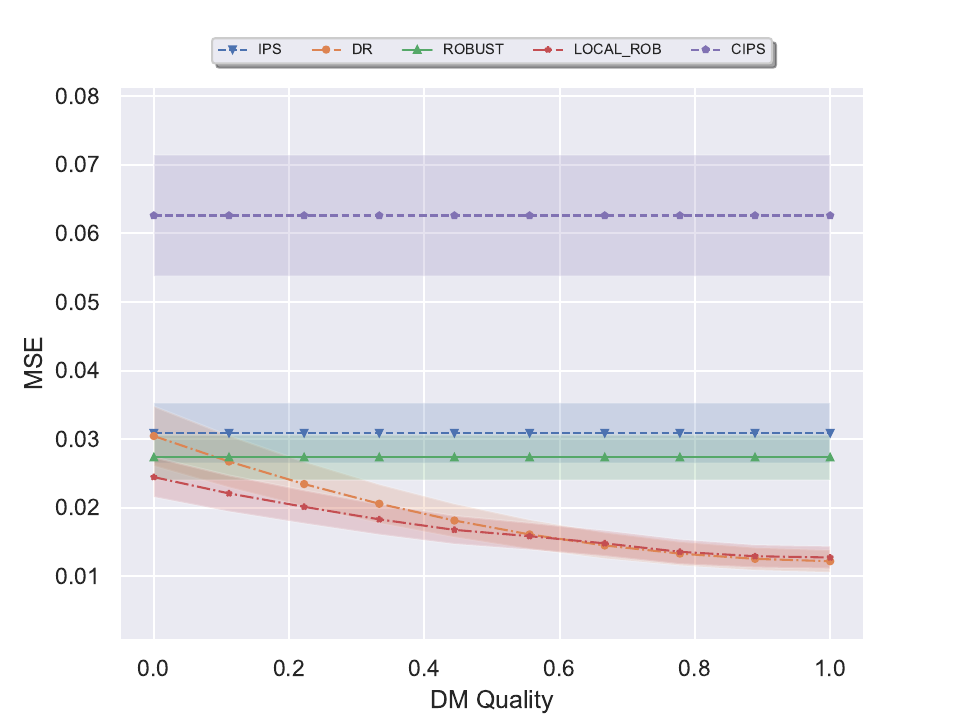}}
	\subfloat[n = 500]{\includegraphics[width=0.5\textwidth]{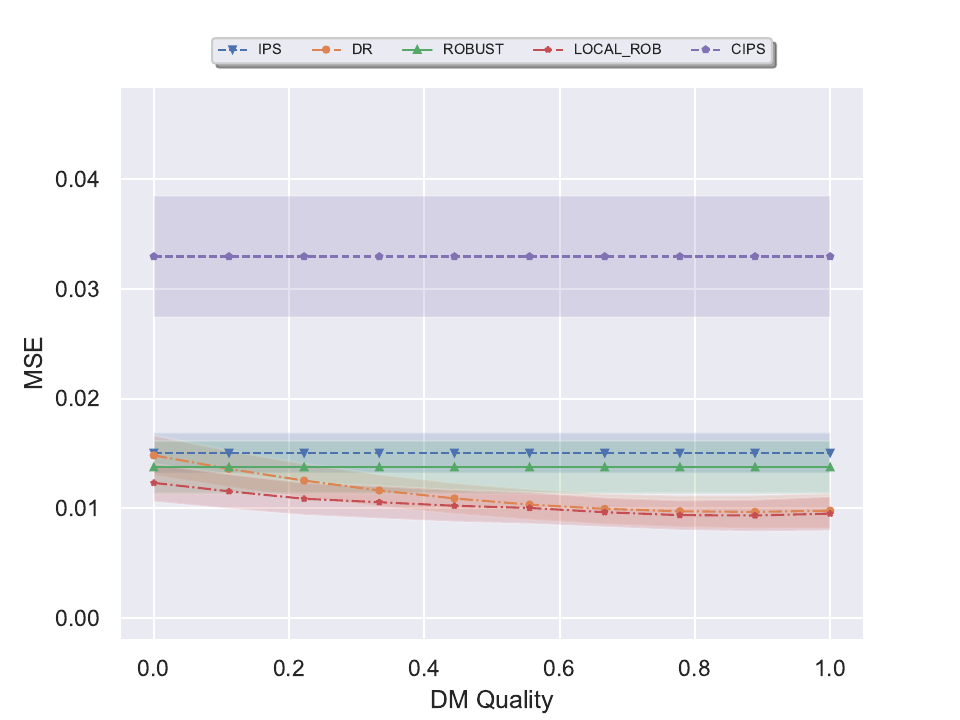}}
	\caption{Loss function evaluation on synthetic data. We use different dataset sizes of 50, 100, 200, and 500. Robust estimator has a better policy evaluation performance than IPS and CIPS estimator here. When the demand estimation is poor, DR estimator has a similar performance to IPS estimator. When the direct method estimation is accurate, DR outperforms IPS and Robust estimators. Locally Robust estimator has the best evaluation performance with varying direct method estimation quality across dataset sizes.}
	\label{fig:eva_datasize}
\end{figure}

We show results in \Cref{fig:eva_datasize} where the x-axis represents the demand estimation quality $\alpha$ and the y-axis represents the mean squared error in estimation for dataset  sizes of 50, 100, 200, and 500. 
It is apparent from \Cref{fig:eva_datasize} that only DR and Locally Robust estimator are dependent on the demand estimation quality $\alpha$.  
When the demand estimate is not accurate, we observe that DR results in inaccurate loss function estimation. Based on our previous analysis, we conjecture that this is because the estimator has a higher variance than some other estimators. 
When demand estimation improves, DR estimates also improve. 
Unlike the DR method, IPS and Robust methods do not require the direct method (demand) estimation and thus are not affected by the direct method's estimation quality. 
In these experiments, the Robust method has a lower MSE (is more accurate) than the IPS method for datasets with various sizes.\footnote{We acknowledge that these results are dependent on the valuation distribution as discussed in Section \ref{sec:ipw_relationship} and that IPS performs well when the selling probabilities are very low. We provide experimental evidence of this in \Cref{sec:ipsvarying}} The gap becomes smaller as the dataset size increases. This is expected since all methods are unbiased, more samples will also decrease the estimation variance and all estimators will reach the unbiased estimation given infinite samples. 
The Locally Robust estimator is generally able to achieve the best of both Robust and DR methods in our experiments by selecting the uncertainty set width $c$ via cross-validation to minimize empirical variance on the observational data. Thus, the Locally Robust estimator may be thought of as an alternative that is safer than the widely used DR baseline in policy evaluation. 

\begin{figure}[!t]
	\centering
	\subfloat[n = 50]{\includegraphics[width=0.5\textwidth]{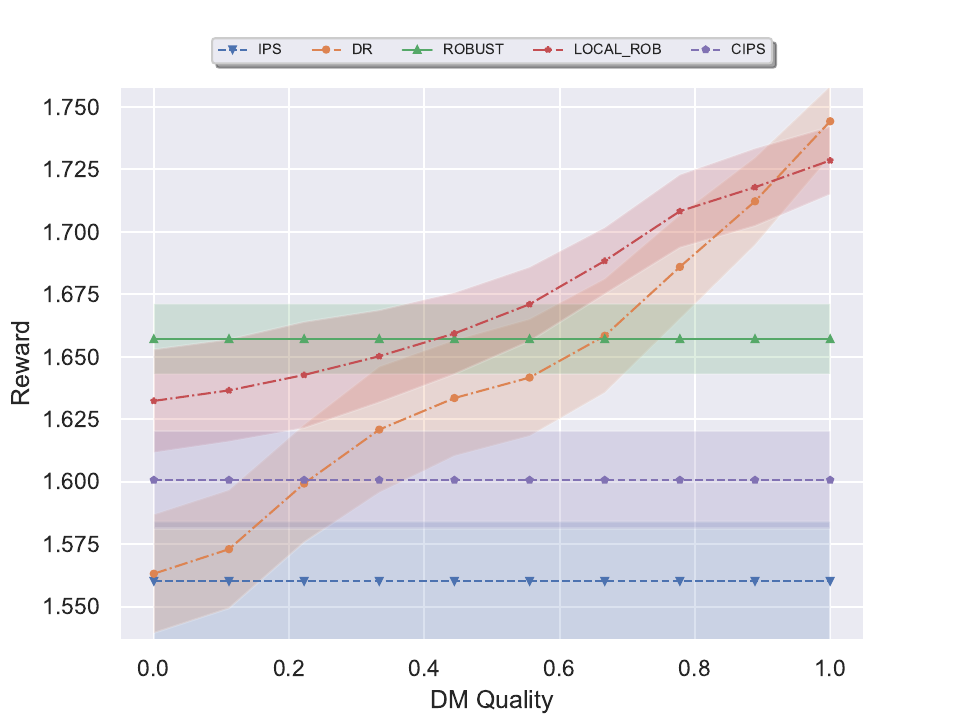}}
	\subfloat[n = 100]{\includegraphics[width=0.5\textwidth]{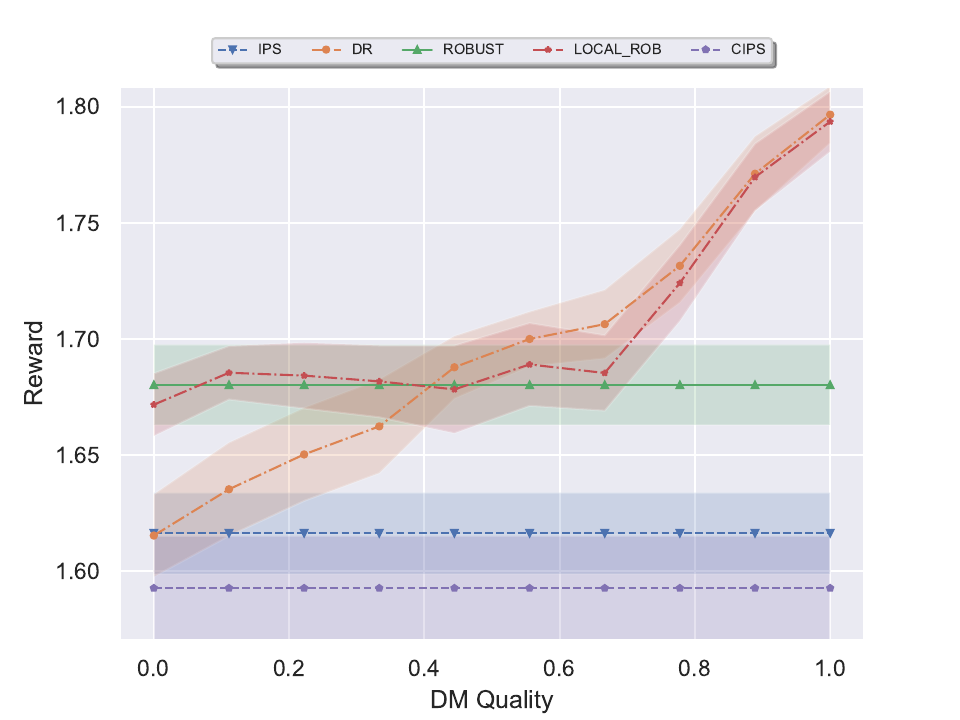}}\\
	\subfloat[n = 200]{\includegraphics[width=0.5\textwidth]{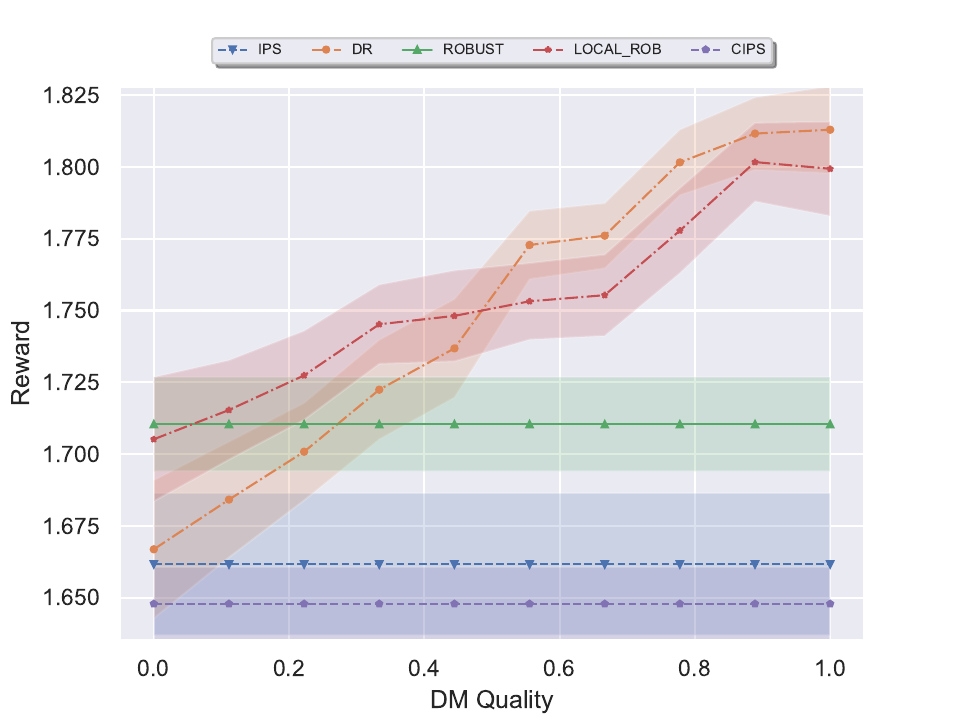}}
	\subfloat[n = 500]{\includegraphics[width=0.5\textwidth]{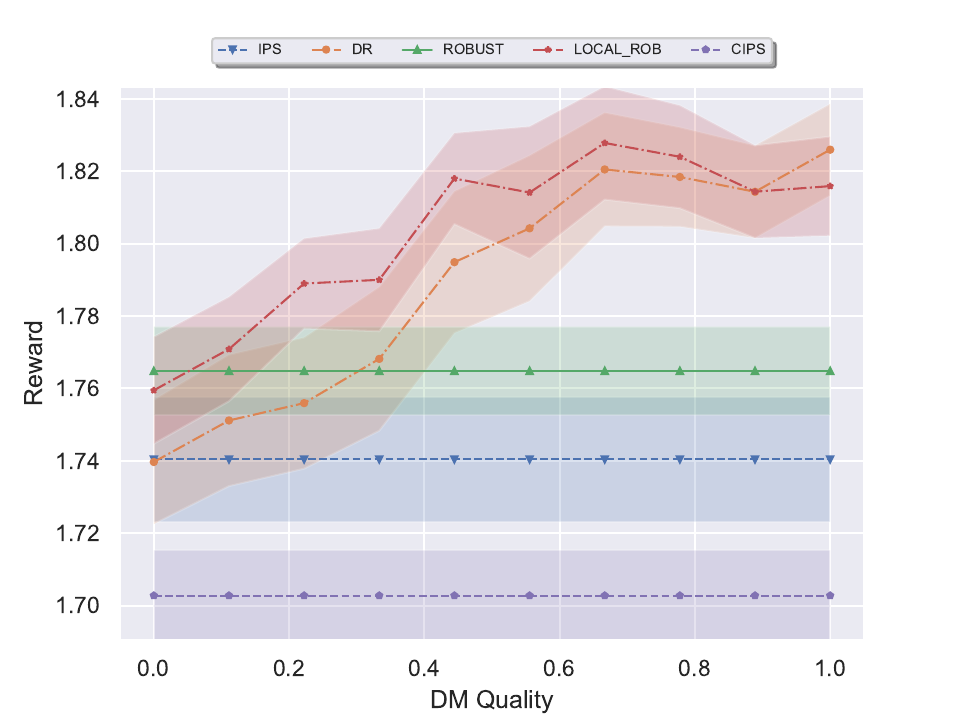}}
	\caption{Loss function optimization on synthetic data. We use different dataset sizes of 50, 100, 200, and 500. Robust estimator has a better policy optimization performance than IPS and CIPS estimator here. When the demand estimation is poor, DR estimator has a similar performance to IPS estimator. When the direct method estimation is accurate, DR outperforms IPS and Robust estimators. Locally Robust estimator has the best optimization performance with varying direct method estimation quality across dataset sizes.}
	\label{fig:learn_datasize}
\end{figure}

For the loss function optimization task, after the observational data are created, we determine an optimal pricing policy by optimizing $\max_\theta \frac{1}{n} \sum_{i=1}^nl_{\tilde{Y}}(\pi_\theta,\tilde{Y}_i)$, where $\theta$ parameterizes the desired policy. The policy class is chosen to be a logistic policy class with a softmax layer to normalize output probabilities, also known as Conditional Random Field~\citep{lafferty2001conditional}. The trained policy is evaluated on a test set of size 10000, and we report the average reward across 20 repetitions. The results are reported in \Cref{fig:learn_datasize} with the training set sizes of 50, 100, 200, and 500. Similar to the loss function evaluation results, the policy returned by the DR estimator achieves lower revenue when the plug-in estimate for the demand is less accurate. As the demand estimation quality improves, the revenue obtained by the policy optimizing the DR estimator improves. The Robust and IPS methods are not dependent on the plug-in estimator, but the Robust method is able to outperform the IPS method and is better than or competitive with all other algorithms at all accuracy levels. The Locally Robust performance is between the Robust and the DR estimators. We observe that the optimization results are different from the evaluation results. In particular, although the Locally Robust estimator is always able to evaluate the policy the most accurately, this does not always correspond to achieving the highest revenue in the optimization setting. We conjecture that this is due to the suboptimal selection of the width of the uncertainty set $c$ for the optimization task, which rather than being a constant should depend on $X$ and the price $p$, but this is computationally challenging.

Nevertheless, we still observe that the methods we propose (Locally Robust, Robust) perform very well compared to DR, one of the most-used policy estimators in practice. Compared to DR, the Locally Robust estimator is relatively more robust when the demand estimator is of poor quality. 
 
 \subsubsection{Loss Function Evaluation and Optimization with Direct Method Demand Estimator}
\label{sec:demand_estimated}

In the previous section, we observed that when the demand estimation quality is poor, the DR estimator has a suboptimal performance compared to the Robust and Locally Robust estimators. Here we provide an example that shows demand estimation can be challenging using modern machine learning techniques. 
We use the direct method of demand estimation where we fit a model to predict the purchase probability as a function of X for each price using gradient boosting trees in the scikit-learn package \citep{friedman2001greedy}. This estimator is also known as the S-learner, value function, or direct comparison in the causal inference literature \citep{shalit2017estimating,kunzel2019metalearners}. This predict-then-optimize procedure is also commonly used in contextual pricing (e.g., \cite{chen2015statistical}). The simulation setup is the same as used previously. %

For policy evaluation, the results are shown in \Cref{tab:eva_datasize} with dataset sizes of 50, 100, 200, and 500. In these experiments, the Robust estimator generally performs the best against other baselines, while the Locally Robust estimator also performs well. We note that the gradient boost tree (DM) performs worse than all policy-based approaches in policy evaluation, suggesting that this is a challenging setting for demand estimation. 
When the sample size becomes larger, the difference across methods becomes smaller, which is expected since all estimators should reach optimal performance given infinite samples. This is consistent with our intuition that the Robust estimator may offer efficient policy evaluation with limited data. 

\begin{table}[!ht]
    \centering
    \caption{Loss function evaluation on synthetic data with direct method estimation (MSE). Results are averaged over 100 runs. We report the mean and standard error. DM, DR, and Locally Robust use gradient boosting trees to estimate demand from historical data. %
    }   
    \begin{tabular}{cccccccccc}\toprule 
         & CIPS & IPS & Robust & DM & DR & Local Rob \\ \midrule 
        N=50 & 0.342$\pm$0.070 & 0.152$\pm$0.029 & 0.125$\pm$0.026 & 0.338$\pm$0.043&0.329$\pm$0.043 &  \textbf{0.114$\pm$0.023}\\ 
        N=100 & 0.198$\pm$0.044 & 0.082$\pm$0.012 & \textbf{0.077$\pm$0.016} & 0.290$\pm$0.036&0.257$\pm$0.033  & \textbf{0.077$\pm$0.018}\\ 
        N=200 & 0.066$\pm$0.015 & 0.033$\pm$0.007 & 0.029$\pm$0.005 & 0.155$\pm$0.018&0.101$\pm$0.013 &  \textbf{0.026$\pm$0.004}\\ 
        N=500 & 0.025$\pm$0.006 & 0.014$\pm$0.002 & 0.011$\pm$0.002 & 0.081$\pm$0.007&0.028$\pm$0.004 &  \textbf{0.010$\pm$0.002}\\ \bottomrule
    \end{tabular}
    \label{tab:eva_datasize}
\end{table}

With the same setup, we present the results for policy optimization in \Cref{tab:learn_mis}. 
Similarly, the Robust and Locally Robust estimators  generally lead to better revenue maximization outcomes. When the sample size is large, all methods have similar performance, and the difference between the DR and Robust methods becomes insignificant since the direct estimator also improves with more samples. 
While the direct method performs much better in policy optimization than policy evaluation, interestingly, we find the conventional predict-then-optimize approach using machine learning models still performs suboptimally in this case, which suggests our proposed estimators may be a promising alternative for price recommendations.

\begin{table}[!ht]
    \centering
    \caption{Loss function optimization on synthetic data with direct method model. Revenue is shown. Results are averaged across 20 runs. We report the mean and standard error. DM, DR and Locally Robust use gradient boosting trees to estimate demand from historical data. %
    }
    \begin{tabular}{ccccccc}\toprule
          & CIPS & IPS & Robust & DM & DR & Local Rob \\ \midrule 
     N=50 & 1.63$\pm$0.01 & 1.56$\pm$0.02 & \textbf{1.66$\pm$0.01} & 1.62$\pm$0.02 & 1.63$\pm$0.02 & \textbf{1.66$\pm$0.02}\\ 
      N=100 & 1.65$\pm$0.01 & 1.63$\pm$0.02 & \textbf{1.68$\pm$0.02} & 1.66$\pm$0.02 & 1.66$\pm$0.02 & \textbf{1.68$\pm$0.02} \\ 
     N=200 & 1.68$\pm$0.01 & 1.68$\pm$0.02 & 1.71$\pm$0.02 & 1.70$\pm$0.01 & 1.72$\pm$0.01 & \textbf{1.73$\pm$0.01} \\ 
     N=500 & 1.71$\pm$0.01 & 1.75$\pm$0.02 & \textbf{1.77$\pm$0.01} & 1.75$\pm$0.01 & \textbf{1.77$\pm$0.01} & 1.76$\pm$0.02 \\ \bottomrule
    \end{tabular}
    \label{tab:learn_mis}
\end{table}

\subsubsection{IPS and CIPS Estimators with Varying Purchase Probabilities}
\label{sec:ipsvarying}
 
To validate our theoretical analysis from \Cref{sec:IPS_results} and \Cref{sec:CIPS}, that the IPS estimator offers strong performance when the overall purchase probabilities are low and CIPS estimator should perform well when such probabilities are high, we compare the performance of IPS and CIPS estimators by varying underlying purchase probabilities. Following the same synthetic data generation process in \Cref{sec:varying_accuracy}, we subtract and add a constant $c=5$ from the logit of sales probability: when $c<0$, it corresponds to a setting where most items are less likely than in our previous setup to sell at any price; when $c>0$, all items are more likely to sell at any given price. In the experiment, the dataset size is 500. The results of loss function evaluation and optimization are included in \Cref{tab:ips_eva} and \Cref{tab:ips_learn}. We also show the performance of the Robust estimator for comparison. 

When the underlying sale probabilities are relatively low, the IPS estimator starts to perform the best in both policy evaluation and optimization, while CIPS has suboptimal performance. Meanwhile, when the sale probabilities are high, the CIPS estimator's performance increases, and IPS estimator's performance decreases rapidly.  This result confirms  our theoretical analysis and %
provides an interesting insight into how to choose the appropriate reward estimator based on domain knowledge.

\begin{table}[!ht]
    \centering
     \caption{Loss function evaluation for IPS, CIPS, and Robust Methods with varying purchase probabilities. We report MSE of these methods under different settings. %
     }    
    \begin{tabular}{cccc}\toprule 
         & CIPS & IPS & Robust \\ \midrule 
        Low Sale Probability& 0.01866$\pm$0.00481 & \textbf{0.00055$\pm$0.00009} & 0.00488$\pm$0.00117 \\ 
        Medium Sale Probability& 0.025$\pm$0.006 & 0.081$\pm$0.007 & \textbf{0.010$\pm$0.002} \\ 
        High Sale Probability& \textbf{0.011$\pm$0.002} & 0.080$\pm$0.016 & 0.022$\pm$0.005 \\ \bottomrule
    \end{tabular}
     \label{tab:ips_eva}
\end{table}

 \begin{table}[!ht]
    \centering
\caption{Loss function optimization for IPS, CIPS, and Robust Methods with varying purchase probabilities. We report revenues of these methods under different settings. %
}    
    \begin{tabular}{cccc}\toprule
         & CIPS & IPS & Robust \\ \midrule
        Low Sale Probability & 0.114$\pm$0.005 & \textbf{0.188$\pm$0.002} & 0.119$\pm$0.005 \\ 
        Medium Sale Probability& 1.708$\pm$0.012 & 1.751$\pm$0.018 & \textbf{1.769$\pm$0.014} \\ 
        High Sale Probability& \textbf{4.366$\pm$0.029} & 4.002$\pm$0.026 & 4.189$\pm$0.021 \\ \bottomrule
    \end{tabular}
     \label{tab:ips_learn}    
\end{table}
 
\subsection{Case Study: Willingness to Pay for Vaccination}

We use a case study from~\citet{slunge2015willingness}, which was also studied in ~\citet{kallus2021fairness}. This case studies the willingness-to-pay for vaccination against tick-borne encephalitis in Sweden. %
While the dataset does not contain counterfactuals, it is a randomized controlled study, making it favorable for evaluating the algorithms.
Originally this case was designed to study the fairness implications of contextual pricing, but here we look at how different algorithms perform in terms of loss function evaluation and optimization. 
The vaccine for tick-borne encephalitis (TBE) is elective, and the study tries to access determinants of willingness to pay to inform health policy. 
Demand is associated with price and income, as well as individual contextual factors such as age, geographic risk factors, trust, perceptions, and knowledge about tick-borne disease. 
The study was a contingent valuation study that asked individuals about uptake at a  price of 100, 250, 500, 750, or 1000 Swedish krona (SEK) uniformly at random. The study finds that “The current
market price of the TBE vaccine deters a substantial share of at-risk people with low incomes from getting vaccinated.” We remove data points with missing purchase responses, so the resulting dataset has 1151 samples with a binary outcome (buy or not buy).

Since counterfactuals are not available in this dataset and prices were assigned uniformly at random, similar to~\citet{kallus2021more}, we use the Horvitz-Thompson estimator \citep{horvitz1952generalization} to evaluate the policy evaluation and optimization. The Horvitz-Thompson estimator is an IPS-like estimator that uses the propensity score to remove the selection bias in the observational data. %
We set aside 70\% of the data as the test set for more accurate evaluation and vary the size of the training set to examine each method. 

The results on loss function evaluation and optimization which are averaged across 200 runs are reported in \Cref{tab:eva_vac} and \Cref{tab:learn_vac} respectively. The demand estimator (direct method) is fitted using gradient boosting trees. For evaluation, the performance of all methods increases when the dataset sizes increase. The Robust and Locally Robust estimators generally have better performance, and the Robust estimator shows the best performance across different training set sizes. 
For optimization, the performance of all methods increases when the dataset sizes increase. Robust and Locally Robust estimators still show competitive performance against other baselines. 
Interestingly, we find that although CIPS has a relatively poor evaluation performance, it sometimes still performs well in optimization. Empirically, we observe that CIPS tends to recommend higher prices compared to other methods, so it performs well when the optimal price is high, which is the case in this setting.
In general, we still observe that often better evaluation methods can lead to better optimization performance, which is consistent with previous work~\citep{dudik2014doubly}, and  our proposed Locally Robust estimator can achieve stable performance in various scenarios.

\begin{table}[]
    \centering
    \caption{Policy evaluation on Vaccination data with direct method model. %
    }
    \resizebox{\linewidth}{!}{
\begin{tabular}{ccccccc}
    \toprule
         & CIPS & IPS & Robust& DM & DR & Local Rob  \\ \hline
        10\% & 6856.18$\pm$746.95 & 5736.41$\pm$569.76  & \textbf{3944.46$\pm$390.37} & 9615.62$\pm$733.09 & 6979.02$\pm$578.51 & 4082.7$\pm$408.17 \\ 
        20\% & 4828.03$\pm$453.44 & 3526.38$\pm$365.72 & \textbf{2673.41$\pm$267.11} & 5149.21$\pm$465.47 & 4158.48$\pm$396.23 & 3061.08$\pm$316.68  \\ 
        30\% & 3325.13$\pm$308.54 & 2592.82$\pm$267.35 & \textbf{1971.47$\pm$205.03}& 4183.96$\pm$382.74 & 3505.03$\pm$330.74 & 2216.92$\pm$225.29  \\ \bottomrule
    \end{tabular}}
    \label{tab:eva_vac}
\end{table}

\begin{table}[]
    \centering
        \caption{Loss function optimization on Vaccination data with direct method model. %
        }
\resizebox{\linewidth}{!}{
\begin{tabular}{ccccccc}
    \toprule
          & CIPS & IPS & Robust & DM & DR & Local Rob \\ \hline
       10\%& 256.9$\pm$9.37 & 246.43$\pm$9.72 & 259.09$\pm$7.25& 213.6$\pm$7.02 & 256.57$\pm$6.82 & \textbf{265.5$\pm$8.15}   \\
       20\%& 271.43$\pm$6.3 & 262.99$\pm$6.95 & \textbf{274.43$\pm$8.25}& 250.06$\pm$7.89 & 266.23$\pm$9.27 & 269.24$\pm$9.04  \\ 
       30\%& 278$\pm$8.64 & 265.18$\pm$6.19  & 276.14$\pm$7.57& 259.98$\pm$7.36 & 273.9$\pm$7.21 & \textbf{282.87$\pm$7.35} \\ \bottomrule
    \end{tabular}}
    \label{tab:learn_vac}
\end{table}

\section{Conclusion and Future Work}

In this paper, we propose loss functions for discrete contextual pricing with observational data. Since the true valuations of customers are rarely known, we adopt theory from learning from noisy supervision to transform observed purchase outcomes to a suitable loss used for true valuations. This allows us to derive a class of unbiased loss functions for pricing, rather than following an indirect predict-then-optimize approach where an intermediate demand function is learned first. Rather than disposing of the estimated demand function entirely, we show that demand learning is helpful in our framework when it is accurate. In this case, we derive the minimum variance loss function, which uses the demand estimation as a plug-in estimate. However, in challenging real-world environments, demand estimation may be inaccurate for reasons like model misspecification, overfitting, and poor uncertainty calibration when using modern machine learning models. In this case, we propose the robust loss function that performs well when demand learning is difficult. %
Using both synthetic and real-world datasets, we provide experimental evidence  that these approaches work well for contextual pricing.

Interestingly, we find that IPS and DR estimators from causal inference literature are a special case of our proposed unbiased loss functions. Specifically, the DR estimator is equivalent to the minimum variance loss function when using a plug-in estimator and applied to contextual pricing. We also show theoretically and empirically that when most items do not sell (i.e., have very low purchase probability), the IPS method is a minimum variance estimator, which implies that practitioners can use their domain knowledge about the demand to inform model selection. To the best of our knowledge, these connections between causal inference, learning from noisy supervision techniques, and contextual pricing are unknown in these communities. 

Our primary focus in this paper is on unbiased loss estimation, but we think a promising area for future work is in investigating how to optimize this loss function to find optimal pricing policies. While more accurate loss estimation generally leads to improved pricing policies, there are additional optimization considerations that can be incorporated into the loss functions we propose. This includes formulating convex surrogates \citep{pires2013cost} and adding restrictions on the variance of the pricing policy \citep{swaminathan2015counterfactual,bertsimas2018optimization}.  Our loss transformation is adapted from the learning from noisy labels literature, which more naturally applies to discrete price levels. It is less clear how to adapt this framework to the continuous pricing setting, which is a generalization we leave for future work. Finally, we study a fundamental yet simple pricing problem of proposing and evaluating loss functions for pricing. Future work may also incorporate additional pricing considerations such as inventory, finite selling periods, and cross-product effects from multiple products.

\bla

\bibliographystyle{plainnat} %
\bibliography{bibliography.bib} %

 \appendix 
\section{Proof of Lemma \ref{transformation_of_dist}}

\proof{Proof of Lemma \ref{transformation_of_dist}:} 

The $i^{th}$ row of $\bm{T}$ is represented as $\bm{T}_{i*}$, while $i^{th}$ column is $\bm{T}_{*i}$.

For $i<=m:$ 
\begin{align}
    f_{\tilde{Y}_i} &=\mathbb{P}(P=p_i,Y(p_i)=1|X)\\
    &=\mathbb{P}(P=p_i, V \geq p_i |X)\\
    &=\mathbb{P}(P=p_i|X) \mathbb{P}(\cap_{j=i}^m (V = p_j) |X) \\
    &= \pi_0(p_i|X) \sum_{j=i}^m f_{V_j}\\
    &= \bm{T}_{i*} \bm{f_{V}}.
\end{align} 

The second equality follows from the monotonicity of the customers' price response, while the third equality follows from the ignorabilty assumption.

Similarly, for $i>m$, define $k=i-m$ 
\begin{align}
    f_{\tilde{Y}_i}&=\mathbb{P}(P=p_k,Y(p_k)=0|X)\\
    &=\mathbb{P}(P=p_k, V < p_k |X)\\
    &=\mathbb{P}(P=p_k|X) \mathbb{P}(\cap_{j=0}^k V = p_j |X)\\
    &= \pi_0(p_k|X) \sum_{j=0}^k f_{V_j}\\
    &= \bm{T}_{i*} \bm{f_V} \Halmos
\end{align}
\endproof

\section{Proof of Lemma \ref{min_var_soln}}
\label{min_var_proof}
\proof{Proof of Lemma \ref{min_var_soln}:} 

We can prove Lemma \ref{min_var_soln} by verifying that $\bm{R}_{MV}$ verifies the KKT conditions. First we decompose the variance term into a two parts, one of which is independent of $\bm{R}$, therefore simplifying the KKT conditions. First we rearrange the objective into a more useful form:

\begin{align}
\bm{l_V}^{\pi'}\bm{R} (\mathrm{diag}(\bm{f_{\tilde{Y}}})- \bm{f_{\tilde{Y}}}\bm{f_{\tilde{Y}}}') \bm{R}'\bm{l_V}^{\pi} &=\bm{l_V}^{\pi'}\bm{R} \mathrm{diag}(\bm{f_{\tilde{Y}}}) \bm{R}'\bm{l_V}^{\pi}-  \bm{l_V}^{\pi'}\bm{R} \bm{f_{\tilde{Y}}}\bm{f_{\tilde{Y}}}' \bm{R}'\bm{l_V}^{\pi} \\
& = \bm{l_V}^{\pi'}\bm{R} \mathrm{diag}(\bm{f_{\tilde{Y}}}) \bm{R}'\bm{l_V}^{\pi}-  \bm{l_V}^{\pi'}\bm{R} \bm{T} \bm{f_V} \bm{f_V}' \bm{T}' \bm{R}'\bm{l_V}^{\pi} \\
&=  \bm{l_V}^{\pi'}\bm{R} \mathrm{diag}(\bm{f_{\tilde{Y}}}) \bm{R}'\bm{l_V}^{\pi}-  \bm{l_V}^{\pi} \bm{f_V} \bm{f_V}' \bm{l_V}^{\pi}
\end{align}
We then differentiate the objective to find the KKT conditions:
\begin{align}
\diffp{}{R} \bm{l_V}^{\pi'}\bm{R} (\mathrm{diag}(\bm{f_{\tilde{Y}}})- \bm{f_{\tilde{Y}}}\bm{f_{\tilde{Y}}}') \bm{R}'\bm{l_V}^{\pi} &= \diffp{}{R} \mathrm{tr}(\bm{l_V}^{\pi'}\bm{R} \mathrm{diag}(\bm{f_{\tilde{Y}}}) \bm{R}'\bm{l_V}^{\pi}) \\
&= \diffp{}{R} \mathrm{tr}(\bm{l_V}^{\pi}\bm{l_V}^{\pi'}\bm{R} \mathrm{diag}(\bm{f_{\tilde{Y}}}) \bm{R}')\\
&= 2\bm{l_V}^{\pi}\bm{l_V}^{\pi'}\bm{R} \mathrm{diag}(\bm{f_{\tilde{Y}}})
\end{align}
KKT conditions:
\begin{align}
\diffp{}{R} \bm{l_V}^{\pi'}\bm{R} \mathrm{diag}(\bm{f_{\tilde{Y}}}) \bm{R}'\bm{l_V}^{\pi} + \bm{\Lambda} \diffp{RT}{R}  &= 0  \\
\bm{R}\bm{T}&=\bm{I} \\ 
\implies ~~ 2\bm{l_V}^{\pi}\bm{l_V}^{\pi'}\bm{R} \mathrm{diag}(\bm{f_{\tilde{Y}}}) +  \bm{\Lambda} \bm{T}'  &= 0  \label{eq:KKT_proof_eq_1} \\
\bm{R}\bm{T}&=\bm{I} \label{eq:KKT_idenity}
\end{align}

Suppose $\bm{R} =  (\bm{T}'\mathrm{diag} (\bm{f_{\tilde{Y}}})^{-1}\bm{T})^{-1}\bm{T}' \mathrm{diag}(\bm{f_{\tilde{Y}}})^{-1}$ and $ \bm{\Lambda} = 2\bm{l_V}^{\pi}\bm{l_V}^{\pi'} (\bm{T}'\mathrm{diag}(\bm{f_{\tilde{Y}}})^{-1} \bm{T})^{-1}$. It follows that constraints (\ref{eq:KKT_proof_eq_1}) and (\ref{eq:KKT_idenity}) are satisfied. \Halmos

\endproof

\section{Proof of Theorem \ref{robust_soln}}
\label{robust_proof}

\begin{repeattheorem} $\forall~ i \in \{1,...,m\}$ define:
\begin{equation}
\bm{f_{\tilde{Y}i}^{rob(1)}} =    \begin{cases}
      b_i   & \text{if} ~~  b_i \geq \frac{\pi_0(p_i|X)}{2} \\
      \frac{\pi_0(p_i|X)}{2}    & \text{if} ~~ b_i \leq \frac{\pi_0(p_i|X)}{2} \leq u_i     \\
      u_i   & \text{if} ~~  u_i \leq \frac{\pi_0(p_i|X)}{2} \\
    \end{cases}, ~~
    \bm{f_{\tilde{Y}i}^{rob(0)}} =    \begin{cases}
      \pi_0(p_i|X) -  b_i   & \text{if} ~~  b_i \geq \frac{\pi_0(p_i|X)}{2} \\
      \frac{\pi_0(p_i|X)}{2}    & \text{if} ~~ b_i \leq \frac{\pi_0(p_i|X)}{2} \leq u_i     \\
      \pi_0(p_i|X) -  u_i   & \text{if} ~~  u_i \leq \frac{\pi_0(p_i|X)}{2} \\
    \end{cases}
\end{equation}

Then the solution to (\ref{robust_optimization_problem_start}) is  $\bm{R}^{rob}=  \argmin_R~ z^{rob}(\bm{R}) = (\bm{T}'\mathrm{diag}(\bm{f_{\tilde{Y}}}^{rob})^{-1}\bm{T})^{-1}\bm{T}' \mathrm{diag}(\bm{f_{\tilde{Y}}}^{rob})^{-1} $
 
\end{repeattheorem}

\proof{Proof of Theorem \ref{robust_soln}:}

Define $\mathcal{U}=\{ \bm{f_{\tilde{Y}}} ~|~ \bm{b} \leq \bm{f_{\tilde{Y}}^{(1)}}  \leq \bm{u} ,~~  \bm{f_{\tilde{Y}}^{(1)}} + \bm{f_{\tilde{Y}}^{(0)}} =\bm{\pi_0} \}$ and $\mathcal{R}=\{\bm{R} ~|~ \bm{RT}=\bm{I} \}$ as the feasibility sets of $\bm{f_{\tilde{Y}}}$ and $\bm{R}$, respectively. Robust formulation (\ref{robust_optimization_problem_start}-\ref{robust_optimization_problem_end}) can be reformulated as
\begin{align}
\label{robust_reform_start}
 z^{rob} = \min_{\bm{R} \in \mathcal{R}} & ~z\\
  \text{s.t.}~ &  z \geq \bm{l_V}^{\pi'}\bm{R} (\mathrm{diag}(\bm{f_{\tilde{Y}}})- \bm{f_{\tilde{Y}}}\bm{f_{\tilde{Y}}}') \bm{R}'\bm{l_V}^{\pi}  \qquad \forall ~ \bm{f_{\tilde{Y}}} \in \mathcal{U} \label{robust_reform_end}
 \end{align}
 Note that (\ref{robust_reform_end}) has an infinite number of constraints. Suppose we select any  $\bm{\underline{f}_{\tilde{Y}}} \in \mathcal{U}$, we can look at a formulation with only one constraint:
 \begin{align}
 \label{lower_bound_form_start}
 \mathrm{Let ~ }  \underbar z(\bm{\underline{f}_{\tilde{Y}}}) =  \min_{\bm{R} \in \mathcal{R}}  & ~z\\
  \text{s.t.}~ &  z \geq \bm{l_V}^{\pi'}\bm{R} (\mathrm{diag}(\bm{\underline{f}_{\tilde{Y}}})- \bm{\underline{f}_{\tilde{Y}}}\bm{\underline{f}_{\tilde{Y}}}') \bm{R}'\bm{l_V}^{\pi}   \label{lower_bound_form_end}
 \end{align}
 Therefore $ \underbar z(\bm{\underline{f}_{\tilde{Y}}}) \leq  z^{rob}$ since formulation (\ref{lower_bound_form_start}-\ref{lower_bound_form_end}) is a relaxation of formulation (\ref{robust_reform_start}-\ref{robust_reform_end}), since formulation (\ref{lower_bound_form_start}-\ref{lower_bound_form_end}) only has a subset of the constraints.

Furthermore, for any $\bar{\bm{R}} \in \mathcal{R} $, let
 \begin{align}
 \label{upper_bound_form}
  \bar z(\bar{\bm{R}}) = \max_{\bm{f_{\tilde{Y}}} \in \mathcal{U}} z(\bar{\bm{R}}) = \max_{\bm{f_{\tilde{Y}}} \in \mathcal{U}} ~&  \bm{l_V}^{\pi'}\bar{\bm{R}} (\mathrm{diag}(\bm{f_{\tilde{Y}}})- \bm{f_{\tilde{Y}}}\bm{f_{\tilde{Y}}}') \bar{\bm{R}}'\bm{l_V}^{\pi} 
 \end{align}
 Then $\bar z(\bar{\bm{R}}) \geq  z^{rob}$ since $\bar{\bm{R}}$ is a feasible solution to problem (\ref{robust_reform_start}-\ref{robust_reform_end}). Therefore, if there exists a pair $\bar{\bm{R}} \in \mathcal{R}$, $ \bm{\underline{f}_{\tilde{Y}}} \in \mathcal{U}$ such that $\bar z(\bar{\bm{R}})=  \underbar z(\bm{\underline{f}_{\tilde{Y}}})$, then $\bar{\bm{R}}$ is an optimal solution to (\ref{robust_reform_start}-\ref{robust_reform_end}). We will prove that such a pair exists where $\bm{\underline{f}_{\tilde{Y}}}= \bm{f_{\tilde{Y}}}^{rob} , ~~ \bar{\bm{R}}= \bm{R}^{rob}$, as defined in Theorem \ref{robust_soln}.

  \begin{proposition}
\label{min_R_robust_lemma}
$\argmin_{R \in \mathcal{R}}  \underbar z(\bm{f_{\tilde{Y}}}^{rob}) = \bm{R}^{rob}$. 
\end{proposition}

This is an application of Lemma \ref{min_var_soln}.

 \begin{lemma}
\label{max_fv_robust_lemma}
$\argmax_{\bm{f_{\tilde{Y}}} \in \mathcal{U}}  \bar z(\bm{R}^{rob}) = \bm{f_{\tilde{Y}}}^{rob} $
\end{lemma}

 We can prove Lemma \ref{max_fv_robust_lemma} by showing $\bm{f_{\tilde{Y}}}^{rob}$ verifies the KKT conditions from formulation (\ref{upper_bound_form}), by showing the existence of corresponding dual variables.  
 
 \begin{equation}
 \diffp{z(\bm{R}^{rob})}{{\bm{f_{\tilde{Y}}}}}=\bm{R}^{rob'}\bm{l_V}^{\pi} \odot \bm{R}^{rob'}\bm{l_V}^{\pi} - 2\bm{l_V}^{\pi'}\bm{R}^{rob}\bm{f_{\tilde{Y}}} \bm{R}^{rob'}\bm{l_V}^{\pi} \label{derivative_robust}
 \end{equation}

 Where $\odot$ is the Hadamard product, and (\ref{derivative_robust}) follows from \cite{LaueMG2018}. 
 Then the KKT conditions can be written as
 \begin{align}
KKT:~~~  & -\bm{R}^{rob'}\bm{l_V}^{\pi} \odot \bm{R}^{rob'}\bm{l_V}^{\pi} + 2\bm{l_V}^{\pi'}\bm{R}^{rob}\bm{f_{\tilde{Y}}} \bm{R}^{rob'}\bm{l_V}^{\pi}  +  \begin{bmatrix}  \bm{\mu} \\ \bm{\mu}  \end{bmatrix} +  \begin{bmatrix}  \bm{\lambda}\\ \bm{0}  \end{bmatrix}  - \begin{bmatrix}  \bm{\xi} \\ \bm{0}  \end{bmatrix}   = \bm{0} \label{kkt_eq_1} \\
 &\bm{f_{\tilde{Y}}^{(1)}} \leq \bm{u}  ~~ (\bm{\lambda}), ~~  - \bm{f_{\tilde{Y}}^{(1)}}  \leq - \bm{b}  ~~ (\bm{\xi})  \label{kkt_eq_2}  \\ 
&\bm{f_{\tilde{Y}}^{(1)}} + \bm{f_{\tilde{Y}}^{(0)}}  =\bm{\pi_0}  ~~ (\bm{\mu}) \label{kkt_eq_3} \\
& (f_{\tilde{Yi}}^{(1)}- u_i)\lambda_i  =0,~~ (f_{\tilde{Yi}}^{(1)}- b_i)\xi_i  =0 ~~ \forall ~i \in \{1,...,m\} \label{kkt_eq_4} \\
& \bm{\lambda}, \bm{\xi} \geq \bm{0} \label{kkt_eq_5} \
\end{align}

$\bm{f_{\tilde{Y}}}^{rob}$ clearly satisfies (\ref{kkt_eq_2}) and (\ref{kkt_eq_3}). For variables  $f_{\tilde{Yi}}^{rob(1)}$ define the corresponding $ \xi^{rob}_i$ and $ \lambda^{rob}_i$ as follows $\forall ~i \in \{1,...,m\}$: 

\begin{equation}
 \xi^{rob}_i=   \begin{cases}
     0   & \text{if} ~~  f_{\tilde{Yi}}^{(1)} > b_i \\
      \geq 0   & \text{if} ~~  f_{\tilde{Yi}}^{(1)} = b_i \\
    \end{cases} ~~
     \lambda^{rob}_i=   \begin{cases}
     0   & \text{if} ~~  f_{\tilde{Yi}}^{(1)} < u_i \\
      \geq 0   & \text{if} ~~~~  f_{\tilde{Yi}}^{(1)} = u_i 
    \end{cases}
\end{equation}

Therefore, $\bm{\xi}^{rob}, \bm{\lambda}^{rob}$  satisfies the complementary  (\ref{kkt_eq_4}) and non-negativity (\ref{kkt_eq_5}) conditions. To show the existence of dual variables such that the KKT conditions are satisfied for $\bm{f_{\tilde{Y}}}^{rob}$, we require certain conditions on the derivative of the objective (\ref{derivative_robust}). Specifically, if we analyze the derivative for the $i^{th}$ and $(i+m)^{th}$ rows, we have
\begin{align}
    \bm{e_i}'(-\bm{R}^{rob'}\bm{l_V}^{\pi} \odot \bm{R}^{rob'}\bm{l_V}^{\pi} + 2\bm{l_V}^{\pi'}\bm{R}^{rob}\bm{f_{\tilde{Y}}}^{rob}  \bm{R}^{rob'}\bm{l_V}^{\pi})  +  \mu_i + \lambda_i   - \xi_i   = 0 \\
    \bm{e_{i+m}}'(-\bm{R}^{rob'}\bm{l_V}^{\pi} \odot \bm{R}^{rob'}\bm{l_V}^{\pi} + 2\bm{l_V}^{\pi'}\bm{R}^{rob}\bm{f_{\tilde{Y}}}^{rob} \bm{R}^{rob'}\bm{l_V}^{\pi})  +  \mu_i = 0 
\end{align}

Therefore, if the following conditions are satisfied, there exist variables $ \mu_i, \lambda_i, \xi_i$ such that the KKT conditions are satisfied:

\begin{equation}
\label{KKT_manipulation}
(\bm{e_i}'-\bm{e_{i+m}'})(\bm{R}^{rob'}\bm{l_V}^{\pi} \odot \bm{R}^{rob'}\bm{l_V}^{\pi} - 2\bm{l_V}^{\pi'}\bm{R}^{rob}\bm{f_{\tilde{Y}}}^{rob} \bm{R}^{rob'}\bm{l_V}^{\pi}) =   \begin{cases}
     \geq  0   & \text{if} ~~  f_{\tilde{Yi}}^{rob(1)} = b_i \\
      = 0   & \text{if} ~~  f_{\tilde{Yi}}^{rob(1)} = \frac{\pi_0(p_i|X)}{2} \\
      \leq 0   & \text{if} ~~  f_{\tilde{Yi}}^{rob(1)} = u_i 
    \end{cases}
\end{equation}

Before we prove this, we need to introduce some additional notation. Define:

 \[ \bm{U} = 
\begin{bmatrix} 
 0 & 1 & 1   & \cdots &  1 & 1 \\
 0 & 0 & 1  & \cdots &  1 & 1\\
  \vdots & \vdots & \vdots  & \ddots & \vdots & \vdots   \\
 0 & 0  & 0 & \cdots &  1 & 1 \\
 0 & 0 &   0 & \cdots &  0& 1 \\
\end{bmatrix}
\qquad \bm{L} = 
\begin{bmatrix} 
 1 & 0 & 0   & \cdots &  0 & 0 \\
 1 & 1 & 0  & \cdots &  0 & 0\\
  \vdots & \vdots & \vdots  & \ddots & \vdots & \vdots  \\
 1 & 1  & 1 & \cdots &  0 & 0 \\
 1 & 1 &   1 & \cdots &  1& 0 \\
\end{bmatrix}
 \]
 
 \noindent with $\bm{U},\bm{L} \in \{0,1\}^{m \times (m+1)}$ such that $\bm{T}'= \begin{bmatrix} \bm{U}' & \bm{L}'  \end{bmatrix}  \mathrm{diag}(\bm{\pi_0},\bm{\pi_0})  $. Also define $\tilde{e_j}=\bm{U}'\bm{e_j}= \bm{e}-\bm{L}'\bm{e_j} =\begin{bmatrix} 0,...,0,1,1,...,1 \end{bmatrix}$. Also define $\bm{T}^\dagger=(\bm{T}'\mathrm{diag}(\bm{f_{\tilde{Y}}}^{rob})^{-1}\bm{T})^{-1}$. It follows that

\begin{align}
    (\bm{e_i}'-\bm{e_{i+m}'})&(\bm{R}^{rob'}\bm{l_V}^{\pi} \odot \bm{R}^{rob'}\bm{l_V}^{\pi} - 2\bm{l_V}^{\pi'}\bm{R}^{rob}\bm{f_{\tilde{Y}}}^{rob} \bm{R}^{rob'}\bm{l_V}^{\pi})  \\
        &=\bm{l_V}^{\pi'} (\bm{R}^{rob} \mathrm{diag}(\bm{e_i},-\bm{e_i}) \bm{R}^{rob'}\bm{l_V}^{\pi} - 2\bm{l_V}^{\pi'}\bm{R}^{rob}\bm{f_{\tilde{Y}}}^{rob}(\bm{e_i}',-\bm{e_i}') \bm{R}^{rob'}\bm{l_V}^{\pi})\\
        &= \bm{l_V}^{\pi'} \bm{T}^\dagger [ \bm{T}' \mathrm{diag}(\bm{f_{\tilde{Y}}}^{rob})^{-1} \mathrm{diag}(\bm{e_i},-\bm{e_i})      \mathrm{diag}(\bm{f_{\tilde{Y}}}^{rob})^{-1} \bm{T} \\  & \qquad \qquad \qquad  -2 \bm{T}' \mathrm{diag}(\bm{f_{\tilde{Y}}}^{rob})^{-1} \bm{f_{\tilde{Y}}}^{rob}(\bm{e_i}',-\bm{e_i}')     \mathrm{diag}(\bm{f_{\tilde{Y}}}^{rob})^{-1} \bm{T}  ]\bm{T}^{\dagger'} \bm{l_V}^{\pi} \\
        &= \bm{l_V}^{\pi'} \bm{T}^\dagger [ \begin{bmatrix} \bm{U}' & \bm{L}'  \end{bmatrix}  \mathrm{diag}(\bm{\pi_0},\bm{\pi_0})  \mathrm{diag}(\bm{f_{\tilde{Y}}}^{rob})^{-1} \mathrm{diag}(\bm{e_i},-\bm{e_i})      \mathrm{diag}(\bm{f_{\tilde{Y}}}^{rob})^{-1}   \mathrm{diag}(\bm{\pi_0},\bm{\pi_0}) \begin{bmatrix}\bm{U} \\ \bm{L}  \end{bmatrix}    \\  & \qquad \qquad \qquad  -2 \bm{T}' \mathrm{diag}(\bm{f_{\tilde{Y}}}^{rob})^{-1} \bm{f_{\tilde{Y}}}^{rob}(\bm{e_i}',-\bm{e_i}')     \mathrm{diag}(\bm{f_{\tilde{Y}}}^{rob})^{-1} \mathrm{diag}(\bm{\pi_0},\bm{\pi_0}) \begin{bmatrix}\bm{U} \\ \bm{L}  \end{bmatrix} ]\bm{T}^{\dagger'} \bm{l_V}^{\pi} \\
        &= \bm{l_V}^{\pi'} \bm{T}^\dagger [ \begin{bmatrix} \bm{U}' & \bm{L}'  \end{bmatrix}   \mathrm{diag}\Big(\left(\frac{\pi_{0i}}{f_{\tilde{Y}i}^{rob}} \right)^2\bm{e_i},-\left(\frac{\pi_{0i}}{\pi_{0i}-f_{\tilde{Y}i}^{rob}} \right)^2\bm{e_i}\Big)   \begin{bmatrix}\bm{U} \\ \bm{L}  \end{bmatrix}    \\  & \qquad \qquad \qquad \qquad  -2 \bm{T}'\bm{e} \Big(\left(\frac{\pi_{0i}}{f_{\tilde{Y}i}^{rob}} \right)\bm{e_i}',-\left(\frac{\pi_{0i}}{\pi_{0i}-f_{\tilde{Y}i}^{rob}} \right)\bm{e_i}'\Big) \begin{bmatrix}\bm{U} \\ \bm{L}  \end{bmatrix} ]\bm{T}^{\dagger'} \bm{l_V}^{\pi} \\
        &= \bm{l_V}^{\pi'} \bm{T}^\dagger \Big[ \left(\frac{\pi_{0i}}{f_{\tilde{Y}i}^{rob}} \right)^2 \bm{U}' \bm{e_i} \bm{e_i}' \bm{U}  - \left(\frac{\pi_{0i}}{\pi_{0i}-f_{\tilde{Y}i}^{rob}} \right)^2 \bm{L}' \bm{e_i}  \bm{e_i}' \bm{L}    \\  & \qquad \qquad \qquad \qquad  -2 \bm{e} \Big(\left(\frac{\pi_{0i}}{f_{\tilde{Y}i}^{rob}} \right)\bm{e_i}' \bm{U}  -\left(\frac{\pi_{0i}}{\pi_{0i}-f_{\tilde{Y}i}^{rob}} \right)\bm{e_i}' \bm{L}\Big)  \Big]\bm{T}^{\dagger'} \bm{l_V}^{\pi} \label{Te_comment} \\
        &= \bm{l_V}^{\pi'} \bm{T}^\dagger \Big[ \left(\frac{\pi_{0i}}{f_{\tilde{Y}i}^{rob}} \right)^2 \bm{\tilde{e}}_i \bm{\tilde{e}}_i'  - \left(\frac{\pi_{0i}}{\pi_{0i}-f_{\tilde{Y}i}^{rob}} \right)^2 (\bm{e}-\bm{\tilde{e}}_i) (\bm{e}-\bm{\tilde{e}}_i)'  \\  & \qquad \qquad \qquad \qquad  -2 \bm{e} \Big(\left(\frac{\pi_{0i}}{f_{\tilde{Y}i}^{rob}} \right)\bm{\tilde{e}}_i' -\left(\frac{\pi_{0i}}{\pi_{0i}-f_{\tilde{Y}i}^{rob}} \right)(\bm{e}'-\bm{\tilde{e}}_i') \Big)  \Big]\bm{T}^{\dagger'} \bm{l_V}^{\pi} \\
        &= \bm{l_V}^{\pi'} \bm{T}^\dagger \Big[ \left(\frac{\pi_{0i}}{f_{\tilde{Y}i}^{rob}} \right)^2 \bm{\tilde{e}}_i \bm{\tilde{e}}_i'  - \left(\frac{\pi_{0i}}{\pi_{0i}-f_{\tilde{Y}i}^{rob}} \right)^2 (\bm{e}\bm{e}'-\bm{\tilde{e}}_i\bm{e}-\bm{e}\bm{\tilde{e}}_i'+\bm{\tilde{e}}_i\bm{\tilde{e}}_i')  \\  & \qquad \qquad \qquad \qquad  -2  \left(\frac{\pi_{0i}}{f_{\tilde{Y}i}^{rob}} \right)\bm{e}\bm{\tilde{e}}_i' +2 \left(\frac{\pi_{0i}}{\pi_{0i}-f_{\tilde{Y}i}^{rob}} \right)(\bm{e}\bm{e}'-\bm{e}\bm{\tilde{e}}_i')  \Big]\bm{T}^{\dagger'} \bm{l_V}^{\pi} \\
        &= \bm{l_V}^{\pi'} \bm{T}^\dagger \Big[ \bm{\tilde{e}}_i\bm{\tilde{e}}_i' \big( \left(\frac{\pi_{0i}}{f_{\tilde{Y}i}^{rob}} \right)^2  - \left(\frac{\pi_{0i}}{\pi_{0i}-f_{\tilde{Y}i}^{rob}} \right)^2 \big) +  \bm{e}\bm{e}'   \big( 2\left(\frac{\pi_{0i}}{\pi_{0i}-f_{\tilde{Y}i}^{rob}} \right)  - \left(\frac{\pi_{0i}}{\pi_{0i}-f_{\tilde{Y}i}^{rob}} \right)^2 \big)  \\ &   -  \bm{\tilde{e}}_i\bm{\tilde{e}}_i' \big(\left(\frac{\pi_{0i}}{\pi_{0i}-f_{\tilde{Y}i}^{rob}} \right)^2 - 2\left(\frac{\pi_{0i}}{\pi_{0i}-f_{\tilde{Y}i}^{rob}} \right) - 2\left(\frac{\pi_{0i}}{f_{\tilde{Y}i}^{rob}} \right)\big)  +  \bm{\tilde{e}}_i\bm{e}'  \left(\frac{\pi_{0i}}{f_{\tilde{Y}i}^{rob}} \right)^2   \Big]\bm{T}^{\dagger'} \bm{l_V}^{\pi} 
\end{align}

In ($\ref{Te_comment}$), we have used the identity $\bm{T}\bm{e}=\bm{e}$, which is due to the structure of $\bm{T}$ and $\bm{\pi}_0$ being a valid probability distribution. Define $\bm{M}= \bm{\tilde{e}}_i\bm{\tilde{e}}_i' \big( \left(\frac{\pi_{0i}}{f_{\tilde{Y}i}^{rob}} \right)^2  - \left(\frac{\pi_{0i}}{\pi_{0i}-f_{\tilde{Y}i}^{rob}} \right)^2 \big) +  \bm{e}\bm{e}'   \big( 2\left(\frac{\pi_{0i}}{\pi_{0i}-f_{\tilde{Y}i}^{rob}} \right)  - \left(\frac{\pi_{0i}}{\pi_{0i}-f_{\tilde{Y}i}^{rob}} \right)^2 \big)  -  \bm{\tilde{e}}_i\bm{\tilde{e}}_i' \big(\left(\frac{\pi_{0i}}{\pi_{0i}-f_{\tilde{Y}i}^{rob}} \right)^2 - 2\left(\frac{\pi_{0i}}{\pi_{0i}-f_{\tilde{Y}i}^{rob}} \right) - 2\left(\frac{\pi_{0i}}{f_{\tilde{Y}i}^{rob}} \right)\big)  +  \bm{\tilde{e}}_i\bm{e}'  \left(\frac{\pi_{0i}}{f_{\tilde{Y}i}^{rob}} \right)^2$. Rather than focusing on $\bm{l_V}^{\pi'} \bm{T}^\dagger$ (which is difficult to analyze) we show that for any $\bm{z} \in \mathbb{R}^m$, (\ref{KKT_manipulation}) is satisfied:

\begin{equation}
\label{if_positive_def}
\bm{z}'\bm{M}\bm{z} =   \begin{cases}
     \geq  0   & \text{if} ~~  f_{\tilde{Yi}}^{rob(1)} \geq \frac{\pi_0(p_i|X)}{2} ~~ (f_{\tilde{Yi}}^{rob(1)} =b_i) \\
      = 0   & \text{if} ~~  f_{\tilde{Yi}}^{rob(1)} = \frac{\pi_0(p_i|X)}{2} \\
      \leq 0   & \text{if} ~~  f_{\tilde{Yi}}^{rob(1)} \leq \frac{\pi_0(p_i|X)}{2} ~~ (f_{\tilde{Yi}}^{rob(1)} =u_i) 
    \end{cases}
\end{equation}

\begin{align}
    \bm{z}'\bm{M}\bm{z} & = \left(\sum_{j= i}^m  z_j \right)^2 \left( \left(\frac{\pi_{0i}}{f_{\tilde{Y}i}^{rob}} \right)^2  - \left(\frac{\pi_{0i}}{\pi_{0i}-f_{\tilde{Y}i}^{rob}} \right)^2 \right) + \left(\sum_{j= 1}^m  z_j \right)^2  \left( 2\left(\frac{\pi_{0i}}{\pi_{0i}-f_{\tilde{Y}i}^{rob}} \right)  - \left(\frac{\pi_{0i}}{\pi_{0i}-f_{\tilde{Y}i}^{rob}} \right)^2 \right)  \\ 
       &+  \left(\sum_{j= 1}^m  z_j \right) \left(\sum_{j= i}^m  z_j \right) \left(\left(\frac{\pi_{0i}}{\pi_{0i}-f_{\tilde{Y}i}^{rob}} \right)^2 - 2\left(\frac{\pi_{0i}}{\pi_{0i}-f_{\tilde{Y}i}^{rob}} \right) - 2\left(\frac{\pi_{0i}}{f_{\tilde{Y}i}^{rob}} \right)\right) +  \left(\sum_{j= i}^m  z_j \right)\left(\sum_{j= 1}^m  z_j \right)  \left(\frac{\pi_{0i}}{f_{\tilde{Y}i}^{rob}} \right)^2 \\
    & = \left(\sum_{j= i}^m  z_j \right)^2 \left(\left(\frac{\pi_{0i}}{f_{\tilde{Y}i}^{rob}} \right)^2  - 2\left(\frac{\pi_{0i}}{f_{\tilde{Y}i}^{rob}} \right)\right) + \left(\sum_{j= i}^m  z_j \right)\left(\sum_{j= 1}^{i-1}  z_j \right) \left( 2\left(\frac{\pi_{0i}}{\pi_{0i}-f_{\tilde{Y}i}^{rob}} \right)- 2\left(\frac{\pi_{0i}}{f_{\tilde{Y}i}^{rob}} \right)\right) \\
    &  \qquad  \qquad   \qquad   + \left(\sum_{j= 1}^{i-1}  z_j \right)^2  \left( 2\left(\frac{\pi_{0i}}{\pi_{0i}-f_{\tilde{Y}i}^{rob}} \right)  - \left(\frac{\pi_{0i}}{\pi_{0i}-f_{\tilde{Y}i}^{rob}} \right)^2 \right)
\end{align}

Let us make the substitution $\alpha=\frac{\pi_{0i}}{f_{\tilde{Y}i}^{rob}}$ such that $\alpha \in \mathbb{R}, \alpha > 1$, since $f_{\tilde{Y}i}+f_{\tilde{Y}(i+m)}=\pi_{0i}$, and  $f_{\tilde{Y}i},f_{\tilde{Y}(i+m)}>0$ by the overlap assumption.

\begin{align}
    \bm{z}'\bm{M}\bm{z} & = \left(\sum_{j= i}^m  z_j \right)^2 (\alpha^2-2\alpha)+ \left(\sum_{j= i}^m  z_j \right)\left(\sum_{j= 1}^{i-1}  z_j \right) \left( \frac{\alpha}{\alpha-1} -\alpha \right) + \left(\sum_{j= 1}^{i-1}  z_j \right)^2 \left( \frac{2\alpha}{\alpha -1}- \frac{\alpha^2}{(\alpha-1)^2}\right) \\
    & = \alpha(\alpha-2) \left[ \left(\sum_{j= i}^m  z_j \right)^2 -  \left(\sum_{j= i}^m  z_j \right)\left(\sum_{j= 1}^{i-1}  z_j \right) \left( \frac{2}{\alpha-1} \right) + \left(\sum_{j= 1}^{i-1}  z_j \right)^2 \left( \frac{1}{(\alpha-1)^2}\right) \right]\\
    & = \alpha(\alpha-2) \left( \left(\sum_{j= i}^m  z_j \right) -  \left(\sum_{j= 1}^{i-1}  z_j \right) \left( \frac{1}{\alpha-1}\right) \right)^2
\end{align}

It follows that the sign of $\bm{z}'\bm{M}\bm{z}$ depends on whether $\alpha \geq 2$. Through substitution, this corresponds to whether $f_{\tilde{Yi}}^{rob(1)} \geq \frac{\pi_0(p_i)}{2}$ and therefore verifies (\ref{if_positive_def}). \Halmos

\section{Proof of Lemma \ref{IPS_lemma}}
\label{IPS_lemma_proof}
\proof{Proof of Lemma \ref{IPS_lemma}:} 
Using the definition of the valuation loss function from \Cref{def:lv_v}, for $\bm{e}_i'\bm{R}_{IPS}' \bm{l_V}^{\pi},$ corresponding to a sale when a price $p_i$ is prescribed, $P=p_i, Y(p_i)=1$, for $1 \leq i \leq m$, we have
\begin{align}
    \bm{e}_i'\bm{R}_{IPS}' \bm{l_V}^{\pi} &= \frac{1}{\pi_0(p_i|X)}l_V(\bm{\pi}(X),p_i)  - \frac{1}{\pi_0(p_i|X)}l_V(\bm{\pi}(X),p_{i-1}) + l_V(\bm{\pi}(X),p_0) \\
    &= \frac{1}{\pi_0(p_i|X)}\left(-\sum_{j=1}^m  \pi(p_j|X)  p_j  \mathbbm{1} \{ p_j \leq p_i \}\right) - \frac{1}{\pi_0(p_i|X)}\left( -\sum_{j=1}^m  \pi(p_j|X)  p_j  \mathbbm{1} \{ p_j \leq p_{i-1} \} \right) + 0 \\
    &=  - \frac{1}{\pi_0(p_i|X)} \left( \sum_{j=1}^m  \pi(p_j|X)  p_j \left( \mathbbm{1} \{ p_j \leq p_i \} -  \mathbbm{1} \{ p_j \leq p_{i-1} \}\right) \right) \\
    &= - \frac{1}{\pi_0(p_i|X)} \left( \sum_{j=1}^m  \pi(p_j|X)  p_j  \mathbbm{1} \{ p_j = p_i \}  \right) 
\end{align}

 \noindent where we used $l_V(\bm{\pi}(X),p_0)=0$. For $m<i\leq 2m$, corresponding to the no-buy case where $P=p_{i-m}, Y(p_{i-m})=0$, we have that $\bm{e}_i'\bm{R}_{IPS}' \bm{l_V}^{\pi}=l_V(\bm{\pi}(X),p_0)=0$. Since there is only a nonzero value when $Y=1$, this can be rewritten as 
 
 $$\tilde{Y}'\bm{R}_{IPS}' \bm{l_V}^{\pi} =  - \sum_{j=1}^m \dfrac{\pi(p_j|X)} {\pi_0(p_j|X)} p_j  Y   \mathbbm{1}\{P=p_j\} \Halmos $$

\endproof

\section{Proof of Lemma \ref{lemma_IPS_min_var}}
\label{IPS_min_proof}

 \proof{Proof of Lemma \ref{lemma_IPS_min_var}:}

We will prove this by showing that the KKT conditions hold for $\bm{R}_{IPS}$ and $\bm{f}_{\tilde{Y}}=\bm{T}\bm{e_{1}}$. These  KKT conditions are derived in
Lemma \ref{min_var_soln} in Equations (\ref{eq:KKT_proof_eq_1}) and (\ref{eq:KKT_idenity}). 
 \begin{align}
 2\bm{l_V}^{\pi}\bm{l_V}^{\pi'}\bm{R}_{IPS} \mathrm{diag}(\bm{T}\bm{e_{1}}) +  \bm{\Lambda} \bm{T}' &= \bm{0}  \label{KKT1_IPS}\\
\bm{R}_{IPS}\bm{T}   &= \bm{I}
\end{align}

In particular, we will show the KKT conditions hold for $\bm{\Lambda}=-2\bm{l_V}^{\pi}\bm{l_V}^{\pi'}\bm{e}_{1}\bm{e}_{1}'$. As in the previous proof, we will use the identity $\bm{T}'= \begin{bmatrix} \bm{U}' & \bm{L}'  \end{bmatrix}  \mathrm{diag}(\bm{\pi_0},\bm{\pi_0}), ~\bm{U},\bm{L} \in \{0,1\}^{m \times (m+1)}$, where 
 \[ \bm{U} = 
\begin{bmatrix} 
 0 & 1 & 1   & \cdots &  1 & 1 \\
 0 & 0 & 1  & \cdots &  1 & 1\\
  \vdots & \vdots & \vdots  & \ddots & \vdots & \vdots   \\
 0 & 0  & 0 & \cdots &  1 & 1 \\
 0 & 0 &   0 & \cdots &  0& 1 \\
\end{bmatrix}
\qquad \bm{L} = 
\begin{bmatrix} 
 1 & 0 & 0   & \cdots &  0 & 0 \\
 1 & 1 & 0  & \cdots &  0 & 0\\
  \vdots & \vdots & \vdots  & \ddots & \vdots & \vdots  \\
 1 & 1  & 1 & \cdots &  0 & 0 \\
 1 & 1 &   1 & \cdots &  1& 0 \\
\end{bmatrix}
 \]
 
Substituting these expressions leads to
\begin{align*}
\bm{\Lambda} \bm{T}'& = -2\bm{l_V}^{\pi}\bm{l_V}^{\pi'}\bm{e}_{1}\bm{e}_{1}' \begin{bmatrix} \bm{U}', & \bm{L}'  \end{bmatrix}  \mathrm{diag}(\bm{\pi_0},\bm{\pi_0}) \\
&= -2\bm{l_V}^{\pi}\bm{l_V}^{\pi'}\bm{e}_{1}\begin{bmatrix} \bm{0}' , & \bm{e}'  \end{bmatrix}  \mathrm{diag}(\bm{\pi_0},\bm{\pi_0}) \\
&=   -2\bm{l_V}^{\pi}\bm{l_V}^{\pi'}\bm{e}_{1} [ \bm{0}',~ \bm{\pi_0}' ]
\end{align*}

Here the second equality holds because $\bm{e}_{1}'\bm{L}'=\bm{e}'$ and $\bm{e}_{1}'\bm{U}'=\bm{0}'$ by inspection.

We note that $\bm{R}_{IPS}$ can be rewritten as $ [ \bm{H} ,\bm{0}_{m+1 \times m}] \mathrm{diag}(\bm{0},\bm{\pi_0}^{-1}) + \bm{e}_{1}\bm{e}'$ for a matrix $H \in \{0,1\}^{(m+1) \times m}$ defined as follows:

\begin{equation}
\bm{H}= 
\begin{bmatrix} 
 -1 & 0 & 0   & \cdots &  0 & 0  \\
 1 & -1 & 0  & \cdots &  0 & 0 \\
  0 & 1 & -1  & \cdots &  0 & 0\\
  0 & 0 & 1  & \cdots &  0 & 0\\
  \vdots & \vdots & \vdots  & \ddots & \vdots & \vdots  \\
 0 & 0  & 0 & \cdots &  1  & -1& \\
 0 & 0 &   0 & \cdots &  0& 1&\\
\end{bmatrix}
\end{equation}

Furthermore, $\mathrm{diag}(\bm{T}\bm{e_{1}})$ is of the form $\mathrm{diag}(\bm{0,\pi})$. As a result, $  [ \bm{H}, \bm{0}_{m+1 \times m} ] \mathrm{diag}(\bm{\pi_0}^{-1},\bm{0}) \mathrm{diag}(\bm{0,\pi})=\bm{0}_{m+1 \times 2m }$, which in turn means $\bm{R}_{IPS} \mathrm{diag}(\bm{T}\bm{e_1}) = \bm{e}_{1}\bm{e}'\mathrm{diag}(\bm{0,\pi})$. It follows that
\begin{align*}
 2\bm{l_V}^{\pi}\bm{l_V}^{\pi'}\bm{R}_{IPS} \mathrm{diag}(\bm{T}\bm{e_{1}}) & =  2\bm{l_V}^{\pi}\bm{l_V}^{\pi'} \bm{e}_{1}\bm{e}' \mathrm{diag}(\bm{0,\pi}) \\
 & =  2\bm{l_V}^{\pi}\bm{l_V}^{\pi'} \bm{e}_{1}\bm{\pi}' 
\end{align*}
 It follows that the KKT condition (\ref{KKT1_IPS}) is satisfied. 

To show that $\bm{R}_{IPS}\bm{T}  = \bm{I}$, we note that by inspection we have $  \bm{H} \bm{U}  = \bm{I} -\bm{e}_{1}\bm{e}'$ and also that $\bm{e}'\bm{T}=\bm{e}'$, since the columns of $\bm{T}$ contain the probability of each historical action $\bm{\pi}$ and therefore sum to 1. It follows that
\begin{align*}
\bm{R}_{IPS}\bm{T}  &= ([  \bm{H},  \bm{0}_{m+1 \times m}] \mathrm{diag}(\bm{\pi_0}^{-1},\bm{0}) + \bm{e}_{1}\bm{e}') \bm{T}  \\
&= [ \bm{H}, \bm{0}_{m+1 \times m}] \mathrm{diag}(\bm{\pi_0}^{-1},\bm{0}) \mathrm{diag}(\bm{\pi_0},\bm{\pi_0}) \begin{bmatrix} \bm{U} \\ \bm{L}  \end{bmatrix}  + \bm{e}_{1}\bm{e}' \bm{T} \\
&= [  \bm{H}, \bm{0}_{m+1 \times m} ]  \begin{bmatrix} \bm{U} \\ \bm{L}  \end{bmatrix}  + \bm{e}_{1}\bm{e}'\\
&=\bm{I} - \bm{e}_{1}\bm{e}' +\bm{e}_{1}\bm{e}'\\
&=\bm{I} \Halmos
\end{align*}

\section{Proof of Lemma \ref{CIPS_lemma}}
\label{CIPS_lemma_proof}

\proof{Proof of Lemma \ref{CIPS_lemma}:}

The proof of Lemma \ref{CIPS_lemma} is very similar to that of Lemma \ref{IPS_lemma}, but we include it for completeness. Using the definition of the valuation loss function from \Cref{def:lv_v}, we will examine $\bm{e}_i'\bm{R}_{CIPS}' \bm{l_V}^{\pi}$, corresponding to $P=p_{i-m}, Y(p_{i-m})=0$ for $m+1 \leq i \leq 2m$. To ease notation, let us define $i'=i-m$:
\begin{align}
    \bm{e}_i'\bm{R}_{IPS}' \bm{l_V}^{\pi} &=   \frac{1}{\pi_0(p_{i'}|X)}l_V(\bm{\pi}(X),p_{i'-1})  - \frac{1}{\pi_0(p_{i'}|X)}l_V(\bm{\pi}(X),p_{i'}) + l_V(\bm{\pi}(X),p_m) \\
    &=  \frac{1}{\pi_0(p_{i'}|X)}\left( -\sum_{j=1}^m  \pi(p_j|X)  p_j  \mathbbm{1} \{ p_j \leq p_{i'-1} \} \right)-  \frac{1}{\pi_0(p_{i'}|X)}\left(-\sum_{j=1}^m  \pi(p_j|X)  p_j  \mathbbm{1} \{ p_j \leq p_{i'} \}\right) \\  & \qquad \qquad \qquad \qquad - \sum_{j=1}^m  \pi(p_j|X)  p_j   \mathbbm{1} \{ p_j \leq p_{m} \}  \\
    &=  - \frac{1}{\pi_0(p_{i'}|X)} \left( \sum_{j=1}^m  \pi(p_j|X)  p_j \left( \mathbbm{1} \{ p_j \leq p_{i'-1} \} -  \mathbbm{1} \{ p_j \leq p_{i'} \}\right) \right) - \sum_{j=1}^m  \pi(p_j|X)  p_j   \\ 
    &= \frac{1}{\pi_0(p_{i'}|X)} \left( \sum_{j=1}^m  \pi(p_j|X)  p_j \mathbbm{1}\{p_j = p_{i'}\}  \right)  - \sum_{j=1}^m  \pi(p_j|X)  p_j  
\end{align}

 For $1<i\leq m$, corresponding to $P=p_i, Y(p_i)=1$, we have that $\bm{e}_i'\bm{R}_{IPS}' \bm{l_V}^{\pi}=l_V(\bm{\pi}(X),p_m)=- \sum_{j=1}^m  \pi(p_j|X)  p_j $. It follows that 
 
 $$\tilde{Y}'\bm{R}_{IPS}' \bm{l_V}^{\pi} = - \sum_{j=1}^{m} p_j \pi(p_j|X) \left(1-\dfrac{(1-Y) \mathbbm{1}\{P=p_j\} } {\pi_0(p_j|X)} \right)\qquad  \Halmos$$

\endproof

\section{Proof of Lemma \ref{CIPS_corollary}}
\label{CIPS_min_var_proof}

 \proof{Proof of Lemma \ref{CIPS_corollary}:} 
 
The proof is very similar to that of Lemma \ref{lemma_IPS_min_var}, but we provide it in full for completeness. We will prove this by showing that the KKT conditions hold for $\bm{R}_{CIPS}$ and $\bm{f}_{\tilde{Y}}=\bm{T}\bm{e_{m+1}}$. These  KKT conditions are derived in
Lemma \ref{min_var_soln} in Equations (\ref{eq:KKT_proof_eq_1}) and (\ref{eq:KKT_idenity}). 
 \begin{align}
 2\bm{l_V}^{\pi}\bm{l_V}^{\pi'}\bm{R}_{CIPS} \mathrm{diag}(\bm{T}\bm{e_{m+1}}) +  \bm{\Lambda} \bm{T}' &= \bm{0}  \label{KKT1_CIPS}\\
\bm{R}_{CIPS}\bm{T}   &= \bm{I}
\end{align}

In particular, we will show the KKT conditions hold for $\bm{\Lambda}=-2\bm{l_V}^{\pi}\bm{l_V}^{\pi'}\bm{e}_{m+1}\bm{e}_{m+1}'$. As in the previous proof, we will use the identity $\bm{T}'= \begin{bmatrix} \bm{U}' & \bm{L}'  \end{bmatrix}  \mathrm{diag}(\bm{\pi_0},\bm{\pi_0}), ~\bm{U},\bm{L} \in \{0,1\}^{m \times (m+1)},   $ where 
 \[ \bm{U} = 
\begin{bmatrix} 
 0 & 1 & 1   & \cdots &  1 & 1 \\
 0 & 0 & 1  & \cdots &  1 & 1\\
  \vdots & \vdots & \vdots  & \ddots & \vdots & \vdots   \\
 0 & 0  & 0 & \cdots &  1 & 1 \\
 0 & 0 &   0 & \cdots &  0& 1 \\
\end{bmatrix}
\qquad \bm{L} = 
\begin{bmatrix} 
 1 & 0 & 0   & \cdots &  0 & 0 \\
 1 & 1 & 0  & \cdots &  0 & 0\\
  \vdots & \vdots & \vdots  & \ddots & \vdots & \vdots  \\
 1 & 1  & 1 & \cdots &  0 & 0 \\
 1 & 1 &   1 & \cdots &  1& 0 \\
\end{bmatrix}
 \]
 
Substituting these expressions leads to
\begin{align*}
\bm{\Lambda} \bm{T}'& = -2\bm{l_V}^{\pi}\bm{l_V}^{\pi'}\bm{e}_{m+1}\bm{e}_{m+1}' \begin{bmatrix} \bm{U}', & \bm{L}'  \end{bmatrix}  \mathrm{diag}(\bm{\pi_0},\bm{\pi_0}) \\
&= -2\bm{l_V}^{\pi}\bm{l_V}^{\pi'}\bm{e}_{m+1}\begin{bmatrix} \bm{e}', & \bm{0}'  \end{bmatrix}  \mathrm{diag}(\bm{\pi_0},\bm{\pi_0}) \\
&=   -2\bm{l_V}^{\pi}\bm{l_V}^{\pi'}\bm{e}_{m+1} [ \bm{\pi_0}',~ \bm{0}' ]
\end{align*}

Here the second equality holds because $\bm{e}_{m+1}'\bm{U}'=\bm{e}'$ and $\bm{e}_{m+1}'\bm{L}'=\bm{0}'$ by inspection.

We note that $\bm{R}_{CIPS}$ can be rewritten as $ [  \bm{0}_{m+1 \times m}, -\bm{H} ] \mathrm{diag}(\bm{0},\bm{\pi_0}^{-1}) + \bm{e}_{m+1}\bm{e}'$ for a matrix $H \in \{0,1\}^{(m+1) \times m}$ defined as follows:

\begin{equation}
\bm{H}= 
\begin{bmatrix} 
 -1 & 0 & 0   & \cdots &  0 & 0  \\
 1 & -1 & 0  & \cdots &  0 & 0 \\
  0 & 1 & -1  & \cdots &  0 & 0\\
  0 & 0 & 1  & \cdots &  0 & 0\\
  \vdots & \vdots & \vdots  & \ddots & \vdots & \vdots  \\
 0 & 0  & 0 & \cdots &  1  & -1& \\
 0 & 0 &   0 & \cdots &  0& 1&\\
\end{bmatrix}
\end{equation}

Furthermore, $\mathrm{diag}(\bm{T}\bm{e_{m+1}})$ is of the form $\mathrm{diag}(\bm{\pi,0})$.  As a result, $  [  \bm{0}_{m+1 \times m}, -\bm{H} ] \mathrm{diag}(\bm{0},\bm{\pi_0}^{-1}) \mathrm{diag}(\bm{\pi,0})=\bm{0}_{m+1 \times 2m }$, which in turn means $\bm{R}_{CIPS} \mathrm{diag}(\bm{T}\bm{e_1}) = \bm{e}_{m+1}\bm{e}'\mathrm{diag}(\bm{\pi,0})$. It follows that
\begin{align*}
 2\bm{l_V}^{\pi}\bm{l_V}^{\pi'}\bm{R}_{CIPS} \mathrm{diag}(\bm{T}\bm{e_{m+1}}) & =  2\bm{l_V}^{\pi}\bm{l_V}^{\pi'} \bm{e}_{m+1}\bm{e}' \mathrm{diag}(\bm{\pi,0}) \\
 & =  2\bm{l_V}^{\pi}\bm{l_V}^{\pi'} \bm{e}_{m+1}\bm{\pi}' 
\end{align*}
 It follows that the KKT condition (\ref{KKT1_CIPS}) is satisfied. 

To show that $\bm{R}_{CIPS}\bm{T}  = \bm{I}$, we note that by inspection we have $  \bm{H} \bm{L}  = -\bm{I} +\bm{e}_{m+1}\bm{e}'$ and also that $\bm{e}'\bm{T}=\bm{e}'$ since the columns of $\bm{T}$ contain the probability of each historical action $\bm{\pi}$ and therefore sum to 1. It follows that
\begin{align*}
\bm{R}_{CIPS}\bm{T}  &= ([  \bm{0}_{m+1 \times m}, -\bm{H} ] \mathrm{diag}(\bm{0},\bm{\pi_0}^{-1}) + \bm{e}_{m+1}\bm{e}') \bm{T}  \\
&= [  \bm{0}_{m+1 \times m}, -\bm{H} ] \mathrm{diag}(\bm{0},\bm{\pi_0}^{-1}) \mathrm{diag}(\bm{\pi_0},\bm{\pi_0}) \begin{bmatrix} \bm{U} \\ \bm{L}  \end{bmatrix}  + \bm{e}_{m+1}\bm{e}' \bm{T} \\
&= [  \bm{0}_{m+1 \times m}, -\bm{H} ]  \begin{bmatrix} \bm{U} \\ \bm{L}  \end{bmatrix}  + \bm{e}_{m+1}\bm{e}'\\
&=\bm{I} - \bm{e}_{m+1}\bm{e}' +\bm{e}_{m+1}\bm{e}'\\
&=\bm{I} \Halmos
\end{align*}

\section{Proof of Theorem \ref{DR_theorem}}
\label{DR_proof}

 \proof{Proof of Theorem \ref{DR_theorem}:}

We will show that $\bm{R}_{MV}$ can be decomposed into three different matrices, $\bm{R}_{MV}= \bm{R}_{IPS}+ \bm{R}_{DM}-\bm{R}_{DIPS}$. Then we will show that each of the decomposed matrices corresponds to a term in the DR estimator. In particular,  

\begin{align}
[\bm{R}_{IPS}'\bm{l_V}^{\pi}]_j&= \frac{p_j Y_i \pi(p_j|X)} {\pi_0(p_j|X)} \mathbbm{1}\{P_i =p_j \} \label{ips_term} \\
[\bm{R}_{DM}'\bm{l_V}^{\pi}]_j&= \sum_{k=1}^m \hat{\mu}_k \pi(p_k|X) \label{dm_term}  \\ 
[\bm{R}_{DIPS}'\bm{l_V}^{\pi}]_j&=\frac{\hat{\mu}_j \pi(p_j|X) }{\pi_0(p_j|X)} \mathbbm{1}\{P_i=p_j \}  \label{dips_term} 
\end{align}

 We begin by defining these matrices $\bm{R}_{IPS},\bm{R}_{DM},\bm{R}_{DIPS}$ in addition to some expressions that are useful for this definition. We note that the $\bm{R}_{IPS}$ matrix used for this proof is subtly different from that defined in Definition \ref{IPS_matrix_definition}.  Let
 
 \[ \bm{U} = 
\begin{bmatrix} 
 0 & 1 & 1   & \cdots &  1 & 1 \\
 0 & 0 & 1  & \cdots &  1 & 1\\
  \vdots & \vdots & \vdots  & \ddots & \vdots & \vdots   \\
 0 & 0  & 0 & \cdots &  1 & 1 \\
 0 & 0 &   0 & \cdots &  0& 1 \\
\end{bmatrix}
\qquad \bm{L} = 
\begin{bmatrix} 
 1 & 0 & 0   & \cdots &  0 & 0 \\
 1 & 1 & 0  & \cdots &  0 & 0\\
  \vdots & \vdots & \vdots  & \ddots & \vdots & \vdots  \\
 1 & 1  & 1 & \cdots &  0 & 0 \\
 1 & 1 &   1 & \cdots &  1& 0 \\
\end{bmatrix}
\qquad \bm{H} = 
\begin{bmatrix} 
 -1 & 0 & 0   & \cdots &  0 & 0  \\
 1 & -1 & 0  & \cdots &  0 & 0 \\
  0 & 1 & -1  & \cdots &  0 & 0\\
  0 & 0 & 1  & \cdots &  0 & 0\\
  \vdots & \vdots & \vdots  & \ddots & \vdots & \vdots  \\
 0 & 0  & 0 & \cdots &  1  & -1& \\
 0 & 0 &   0 & \cdots &  0& 1&\\
\end{bmatrix}
 \]
 
 \noindent with $\bm{U},\bm{L} \in \{0,1\}^{m \times (m+1)}, \bm{H} \in \{0,1\}^{(m+1) \times m} $. Note that $ \bm{U}\bm{H} = \bm{I}, \bm{L}\bm{H} = -\bm{I}$.  We have that $\bm{R}_{IPS}=  \begin{bmatrix} \bm{H}  \mathrm{diag}(\bm{\pi_0})^{-1},  & \bm{0}_{m \times m} \end{bmatrix}$.  Define:
  \begin{align}
  \bm{\hat{f}_{Y_1}}&=[\bm{\hat{f}_Y}]_{1:m}=[\mathbb{P}(P=p_1,Y(p_1)=1|X), ~\mathbb{P}(P=p_2,Y(p_2)=1|X)..., \mathbb{P}(P=p_m,Y(p_m)=1|X)]    \\
  \bm{\hat{f}_{Y_0}}&=[\bm{\hat{f}_Y}]_{(m+1):2m}=[\mathbb{P}(P=p_1,Y(p_1)=0|X), ~\mathbb{P}(P=p_2,Y(p_2)=0|X)..., \mathbb{P}(P=p_m,Y(p_m)=0|X)]    
 \end{align}
 
Note that $\bm{\hat{f}_{Y_1}}+\bm{\hat{f}_{Y_0}}= \bm{\pi_0}$, since $\hat{f}_{Y_i} = \hat{\mathbb{P}}(P=p_i,Y(p_i)=1|X)$ and $\hat{f}_{Y_{i+m}} = \hat{\mathbb{P}}(P=p_i,Y(p_i)=0|X)$, respectively. Furthermore, $ \frac{\pi_0(p_i|X)}{\hat{f}_{Y_{0_i}}}=\frac{\hat{f}_{Y_{1_i}}}{\hat{f}_{Y_{0_i}}} + 1$, an identity that is used in the proof shortly.

Define $\bm{R}_{DIPS}= \begin{bmatrix}  \bm{H} \mathrm{diag}(\frac{\bm{\hat{f}_{Y_1}}}{\bm{\pi_0}^2}),  & \bm{H} \mathrm{diag}(\frac{\bm{\hat{f}_{Y_1}}}{\bm{\pi_0}^2}) \end{bmatrix} $ and $\bm{R}_{DM}=\bm{\hat{f}_V} \bm{e'}.$  We will now show that these matrices are a valid decomposition of $\bm{R}_{MV}$:

\begin{align}
    \bm{R}_{MV}&-(\bm{R}_{DM} +\bm{R}_{IPS} -\bm{R}_{DIPS}) \\ &= (\bm{T}'\mathrm{diag}(\bm{\hat{f}}_{\tilde{Y}})^{-1}\bm{T})^{-1} (\bm{T}'\mathrm{diag}(\bm{\hat{f}}_{\tilde{Y}})^{-1}\bm{T}) (\bm{R}_{MV}- \bm{R}_{DM} +\bm{R}_{IPS} -\bm{R}_{DIPS})   \\
    &= (\bm{T}'\mathrm{diag}(\bm{\hat{f}}_{\tilde{Y}})^{-1}\bm{T})^{-1} \bm{T}'\mathrm{diag}(\bm{\hat{f}}_{\tilde{Y}})^{-1} (\bm{T}\bm{R}_{MV}- \bm{T} \bm{R}_{DM} +\bm{T} \bm{R}_{IPS} -\bm{T} \bm{R}_{DIPS})  
\end{align}
\begingroup
\addtolength{\jot}{0.5em}
\begin{align}
\bm{T} \bm{R}_{MV} & = \bm{T} \mathrm{diag}(\bm{\hat{f}_Y})^{-1} \bm{T} (\bm{T}'\mathrm{diag}(\bm{\hat{f}}_{\tilde{Y}})^{-1}\bm{T})^{-1} = \bm{I}  \\
\bm{T} \bm{R}_{DM} & = \bm{T} \bm{\hat{f}_V} \bm{e'} = \bm{\hat{f}_Y} \bm{e'} \\
& =  \begin{bmatrix}  \bm{\hat{f}_{Y_1}} \bm{e'}  ,  &  \bm{\hat{f}_{Y_1}} \bm{e'} \\
\bm{\hat{f}_{Y_0}} \bm{e'}  ,  &  \bm{\hat{f}_{Y_0}} \bm{e'}  \end{bmatrix} \\
\bm{T} \bm{R}_{IPS} & = \bm{T} \begin{bmatrix} \bm{H}  \mathrm{diag}(\bm{\pi_0})^{-1},  & \bm{0}_{m \times m} \end{bmatrix} \\
&=  \begin{bmatrix} \mathrm{diag}(\bm{\pi_0})  \bm{U}  \\  \mathrm{diag}(\bm{\pi_0}) \bm{L} \end{bmatrix} \begin{bmatrix} \bm{H}  \mathrm{diag}(\bm{\pi_0})^{-1},  & \bm{0}_{m \times m} \end{bmatrix} \\
&= \begin{bmatrix}  \mathrm{diag}(\bm{\pi_0}) \bm{U} \bm{H} \mathrm{diag}(\bm{\pi_0})^{-1},   &  \bm{0}_{m \times m} \\
\mathrm{diag}(\bm{\pi_0}) \bm{L} \bm{H} \mathrm{diag}(\bm{\pi_0})^{-1}, &  \bm{0}_{m \times m} \end{bmatrix} \\
&= \begin{bmatrix}  \bm{I}, &  \bm{0}_{m \times m} \\ -\bm{I}, &  \bm{0}_{m \times m}  \end{bmatrix} \\
\bm{T} \bm{R}_{DIPS} & = \bm{T} \begin{bmatrix}  \bm{H} \mathrm{diag}(\frac{\bm{\hat{f}_{Y_1}}}{\bm{\pi_0}^2}),  & \bm{H} \mathrm{diag}(\frac{\bm{\hat{f}_{Y_1}}}{\bm{\pi_0}^2}) \end{bmatrix}  \\
&=  \begin{bmatrix} \mathrm{diag}(\bm{\pi_0})  \bm{U}  \\  \mathrm{diag}(\bm{\pi_0}) \bm{L} \end{bmatrix} \begin{bmatrix}  \bm{H} \mathrm{diag}(\frac{\bm{\hat{f}_{Y_1}}}{\bm{\pi_0}^2}),  & \bm{H} \mathrm{diag}(\frac{\bm{\hat{f}_{Y_1}}}{\bm{\pi_0}^2}) \end{bmatrix} \\
&=  \begin{bmatrix} \mathrm{diag}(\bm{\pi_0})  \bm{U} \bm{H} \mathrm{diag}(\frac{\bm{\hat{f}_{Y_1}}}{\bm{\pi_0}^2}), &  \mathrm{diag}(\bm{\pi_0})  \bm{U} \bm{H} \mathrm{diag}(\frac{\bm{\hat{f}_{Y_1}}}{\bm{\pi_0}^2})  \\  \mathrm{diag}(\bm{\pi_0}) \bm{L} \bm{H} \mathrm{diag}(\frac{\bm{\hat{f}_{Y_1}}}{\bm{\pi_0}^2}), & \mathrm{diag}(\bm{\pi_0}) \bm{L} \bm{H} \mathrm{diag}(\frac{\bm{\hat{f}_{Y_1}}}{\bm{\pi_0}^2})  \end{bmatrix} \\
&=  \begin{bmatrix}  \mathrm{diag}(\frac{\bm{\hat{f}_{Y_1}}}{\bm{\pi_0}}), &  \mathrm{diag}(\frac{\bm{\hat{f}_{Y_1}}}{\bm{\pi_0}})  \\  -\mathrm{diag}(\frac{\bm{\hat{f}_{Y_1}}}{\bm{\pi_0}}), & -\mathrm{diag}(\frac{\bm{\hat{f}_{Y_1}}}{\bm{\pi_0}})   \end{bmatrix} 
\end{align}
\endgroup

Therefore,
$\bm{T}\bm{R}_{MV}- \bm{T} \bm{R}_{DM} +\bm{T} \bm{R}_{IPS} -\bm{T} \bm{R}_{DIPS}= \begin{bmatrix} \mathrm{diag}(\frac{\bm{\hat{f}_{Y_1}}}{\bm{\pi_0}}) - \bm{\hat{f}_{Y_1}} \bm{e'},   & \mathrm{diag}(\frac{\bm{\hat{f}_{Y_1}}}{\bm{\pi_0}}) - \bm{\hat{f}_{Y_1}} \bm{e'}  \\ 
\mathrm{diag}(\frac{\bm{\hat{f}_{Y_1}}}{\bm{\pi_0}}) - \bm{\hat{f}_{Y_0}} \bm{e'} +\bm{I} ,&
\mathrm{diag}(\frac{\bm{\hat{f}_{Y_1}}}{\bm{\pi_0}}) - \bm{\hat{f}_{Y_0}} \bm{e'} + \bm{I} \end{bmatrix} $
\begin{align}
\mathrm{Finally,}& ~ \bm{T}'\mathrm{diag}(\bm{\hat{f}}_{\tilde{Y}})^{-1} (\bm{T}\bm{R}_{MV}- \bm{T} \bm{R}_{DM} +\bm{T} \bm{R}_{IPS} -\bm{T} \bm{R}_{DIPS}) \\ 
&= [\bm{U}',\bm{L}'] \mathrm{diag}(\bm{\pi_0},\bm{\pi_0})  \mathrm{diag}(\bm{\hat{f}}_{\tilde{Y}})^{-1} \begin{bmatrix} \mathrm{diag}(\frac{\bm{\hat{f}_{Y_1}}}{\bm{\pi_0}}) - \bm{\hat{f}_{Y_1}} \bm{e'},   & \mathrm{diag}(\frac{\bm{\hat{f}_{Y_1}}}{\bm{\pi_0}}) - \bm{\hat{f}_{Y_1}} \bm{e'}  \\ 
\mathrm{diag}(\frac{\bm{\hat{f}_{Y_1}}}{\bm{\pi_0}}) - \bm{\hat{f}_{Y_0}} \bm{e'} +\bm{I} ,&
\mathrm{diag}(\frac{\bm{\hat{f}_{Y_1}}}{\bm{\pi_0}}) - \bm{\hat{f}_{Y_0}} \bm{e'} + \bm{I} \end{bmatrix} \\
&= -\bm{U}' \mathrm{diag}(\frac{\bm{\pi_0}}{\bm{\hat{f}_{Y_1}}}) \bm{\hat{f}_{Y_1}} \bm{e'} + 
\bm{U}' \mathrm{diag}(\frac{\bm{\pi_0}}{\bm{\hat{f}_{Y_1}}}) \mathrm{diag}(\frac{\bm{\hat{f}_{Y_1}}}{\bm{\pi_0}}) \cdots \\
& \qquad \qquad \qquad \cdots
- \bm{L}' \mathrm{diag}(\frac{\bm{\pi_0}}{\bm{\hat{f}_{Y_0}}}) \bm{\hat{f}_{Y_0}} \bm{e'}  
- \bm{L}' \mathrm{diag}(\frac{\bm{\hat{f}_{Y_1}}}{\bm{\pi_0}})
\mathrm{diag}(\frac{\bm{\pi_0}}{\bm{\hat{f}_{Y_0}}})
- \bm{L}' \mathrm{diag}(\frac{\bm{\pi_0}}{\bm{\hat{f}_{Y_0}}}) \\
&= -\bm{U}' \bm{\pi_0} \bm{e'} + \bm{U}' - \bm{L}' \bm{\pi_0} \bm{e'}  
- \bm{L}' \mathrm{diag}(\frac{\bm{\hat{f}_{Y_1}}}{\bm{\hat{f}_{Y_0}}})
+ \bm{L}' \left( \mathrm{diag}(\frac{\bm{\hat{f}_{Y_1}}}{\bm{\hat{f}_{Y_0}}}) +\bm{I} \right)\\
&= -\bm{U}' \bm{\pi_0} \bm{e'} + \bm{U}' - \bm{L}' \bm{\pi_0} \bm{e'} + \bm{L}' \\
&= (\bm{U}'+\bm{L}')(\bm{I}- \bm{\pi_0} \bm{e}') \\
&= \bm{e}\bm{e}'(\bm{I}- \bm{\pi_0} \bm{e}') \\
&= \bm{e}\bm{e}'- \bm{e}\bm{e}'\bm{\pi_0} \bm{e}' \\
&= \bm{e}\bm{e}'- \bm{e}\bm{e}' \\
&= 0
\end{align}

Therefore, $\bm{R}_{MV}=\bm{R}_{DM} +\bm{R}_{IPS} -\bm{R}_{DIPS}$. We now show that each matrix corresponds to a term in the DR estimator, as defined in Equations (\ref{ips_term} - \ref{dips_term}).
\begin{align}
    [\bm{R}_{DM}' \bm{l_V}^{\pi}]_i &=   - \sum_{i=1}^m \hat{f}_{V_i} \sum_{j=1}^m \pi(p_j|X) p_j \mathbbm{1}\{ p_j \leq p_i \} \\ 
    &=   \sum_{j=1}^m \pi(p_j|X)  p_j (\sum_{i=j}^m \hat{f}_{V_i})\\
    &=  \sum_{j=1}^m \pi(p_j|X)   \hat{\mu}_j 
\end{align}
\begin{align}
    [\bm{R}_{DIPS}' \bm{l_V}^{\pi}]_i &=  \frac{1}{\pi_0(p_i|X)^2} \sum_{j=1}^m \pi(p_j|X)   \hat{f}_{Y_1i}  (p_j \mathbbm{1}\{p_j \leq p_i\} - p_j \mathbbm{1}\{p_j \leq p_{i+1}\} ) \\ 
    &= \frac{1}{\pi_0(p_i|X)^2}  \hat{f}_{Y_{1i}}  p_i \pi(p_i|X)  \\ 
    &= \frac{1}{\pi_0(p_i|X)^2}  \bm{T}_i \bm{\hat{f}_V}  p_i \pi(p_i|X)  \\
    &= \frac{1}{\pi_0(p_i|X)^2}  \pi_0(p_i|X) (\sum_{j=i}^m \hat{f}_{V_j})  p_i \pi(p_i|X)  \\
    &= \frac{ \hat{\mu}_i \pi(p_i|X) } {\pi_0(p_i|X)}
\end{align}
 \endproof

From Lemma \ref{IPS_lemma}, $ [\bm{R}_{IPS}' \bm{l_V}^{\pi}]_i = \frac{Y_i p_i \pi(p_i|X) } {\pi_0(p_i|X)}$. \Halmos
\endproof

 \section{Unknown Historical Pricing Policies}
 \label{unknown_pricing_policy}
 
In the setting where the historic pricing policy is not known, the DR method is known to have advantages over the IPS method and the direct method (also known as predict-then-optimize approach), one of the propensity scores $\hat{\pi}_0$ or direct reward estimate $\hat{\mu}$ needs to be accurate for strong performance. Considering that according to Theorem \ref{DR_theorem}, DR is a special case of the class of loss functions we propose, this gives insight into the behavior if the pricing policy isn't known and the direct method can be beaten. We can define the direct method as $l_{DM}(\pi(X),\tilde{Y})= \sum_{k=1}^m \hat{\mu}(X, p_k) \pi(p_k|X)$. This is a weighted average of the predicted revenues for each price, weighted by the probability that price would be chosen under the policy being evaluated. The predicted revenues could come from a machine learning model and potentially have a bias as previously discussed, defined as $d(X,p_i) = \hat{\mu}(X, p_i)-\mu(X,p_i)$, where $\mu(X,p_i) = p_i \mathbb{E}(Y(X,p_i))$.  We can define a multiplicative error in the historic pricing probability as $\delta(X, p_i)= 1 - \frac{\pi_0(p_i|X)}{\hat{\pi_0}(p_i|X)}$. Note that as $\hat{\pi_0}(p_i|X)$ approaches $\pi_0(p_i|X)$, $\delta(p_i)$ approaches 0. Taking expectation over the policy and data, \cite{dudik2011doubly} have characterized the bias of the off-policy learners as follows:

\begin{theorem} (\cite{dudik2011doubly}):
\begin{align}
\mathbb{E}_{\tilde{Y},X,\pi}[l_{DR}(\pi(X),\tilde{Y})]-\mathbb{E}_{\tilde{Y},X,\pi}[l_V(\pi(X),V)] &= \left| \mathbb{E}_{X,\pi}\left[\sum_{i=1}^m \pi(p_i|X) d(X,p_i) \delta(X, p_i ) \right]\right| \\
\mathbb{E}_{\tilde{Y},X,\pi}[l_{IPS}(\pi(X),\tilde{Y})]-\mathbb{E}_{\tilde{Y},X,\pi}[l_V(\pi(X),V)] &= \left| \mathbb{E}_{X,\pi} \left[\sum_{i=1}^m \pi(p_i|X) \mu(X,p_i) \delta(X,p_i) \right]\right| \\
\mathbb{E}_{\tilde{Y},X,\pi}[l_{DM}(\pi(X),\tilde{Y})]-\mathbb{E}_{\tilde{Y},X,\pi}[l_V(\pi(X),V)] &= \left| \mathbb{E}_{X,\pi} \left[\sum_{i=1}^m \pi(p_i|X) d(X,p_i)\right]\right| 
\end{align}
\end{theorem}

This result suggests that if either $d(p_i)=0$ or $\delta(p_i)=0$, then the loss function will be unbiased for the DR estimator. Furthermore, if $\delta \ll 1 $, then the DR estimator will be significantly less biased than the direct method. Similarly, provided $d \ll \mu$, the DR estimator will be less biased than the IPS estimator. 

\end{document}